\documentclass[review]{elsarticle}
\usepackage[colorlinks,linkcolor=pink]{hyperref}
\usepackage{mathtools}
\usepackage{amsthm}
\usepackage{multirow}
\usepackage{rotating}
\usepackage{adjustbox}
\newcommand{\etal}{\textit{et al}.}
\usepackage{makecell}
\usepackage{amsmath,amssymb,amsfonts,bm}
\usepackage{graphicx}
\usepackage{textcomp}
\usepackage[ruled,vlined,linesnumbered]{algorithm2e}
\usepackage{subfigure}
\usepackage{pifont}
\usepackage{adjustbox}
\usepackage{booktabs}
\usepackage{color}
\usepackage[table,xcdraw]{xcolor}
\usepackage{enumerate}
\newcommand{\keywords}[1]{\textbf{Keywords:} #1}
\journal{Pattern Recognition}

\begin{document}

\begin{frontmatter}

\title{Enhancing Incomplete Multi-modal Brain Tumor Segmentation with Intra-modal Asymmetry and Inter-modal Dependency}

\address[HMS] {Boston Children's Hospital, Harvard Medical School, Boston, MA, USA 02115}
\address[NTU] {School of Computer Science and Engineering, Nanyang Technological University, Singapore 639798}
\address[JUFE] {School of Information Management, Jiangxi University of Finance and Economics, China 330000}
\address[Infocomm] {Institute for Infocomm Research, A*STAR, Singapore 138632}
\address[SZU] {School of Biomedical Engineering, Shenzhen University, China 518060}

\author[HMS]{Weide Liu}
\author[NTU]{Jingwen Hou}
\author[JUFE]{Xiaoyang Zhong}
\author[Infocomm]{Huijing Zhan}
\author[Infocomm]{Jun Cheng}
\author[JUFE]{Yuming Fang}
\author[SZU]{Guanghui Yue}

\cortext[cor1]{ Corresponding author. W. Liu is with Boston Children's Hospital, Harvard Medical School, Boston, MA, USA 02115 (e-mail: weide001@e.ntu.edu.sg).}

\begin{abstract}
Deep learning-based brain tumor segmentation (BTS) models for multi-modal MRI images have seen significant advancements in recent years. However, a common problem in practice is the unavailability of some modalities due to varying scanning protocols and patient conditions, making segmentation from incomplete MRI modalities a challenging issue. Previous methods have attempted to address this by fusing accessible multi-modal features, leveraging attention mechanisms, and synthesizing missing modalities using generative models. However, these methods ignore the intrinsic problems of medical image segmentation, such as the limited availability of training samples, particularly for cases with tumors. Furthermore, these methods require training and deploying a specific model for each subset of missing modalities.
To address these issues, we propose a novel approach that enhances the BTS model from two perspectives. Firstly, we introduce a pre-training stage that generates a diverse pre-training dataset covering a wide range of different combinations of tumor shapes and brain anatomy. Secondly, we propose a post-training stage that enables the model to reconstruct missing modalities in the prediction results when only partial modalities are available. To achieve the pre-training stage, we conceptually decouple the MRI image into two parts: `anatomy' and `tumor'. We pre-train the BTS model using synthesized data generated from the anatomy and tumor parts across different training samples.
For the post-training stage, we introduce a knowledge distillation-based process that enables the model to adapt to partial-modality inputs. This process intentionally removes one modality from the input while encouraging the model to produce the same output as when all modalities are present. This allows the model to reconstruct missing information through the post-training process.
Extensive experiments demonstrate that our proposed method significantly improves the performance over the baseline and achieves new state-of-the-art results on three brain tumor segmentation datasets: BRATS2020, BRATS2018, and BRATS2015. The code is available at \href{https://github.com/ZhongAobo/Asymmetry-BTS}{https://github.com/ZhongAobo/Asymmetry-BTS.}

\end{abstract}

\end{frontmatter}

\section{Introduction}
\begin{figure}[t]
  \centering
    \includegraphics[width=0.7\linewidth]{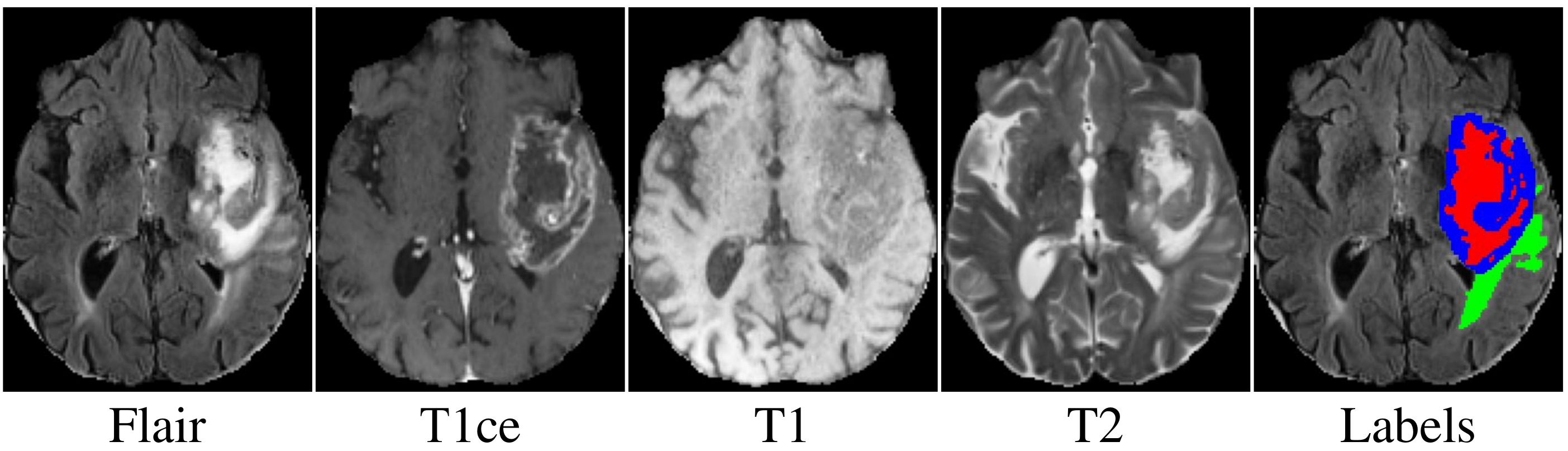}
    \caption{From left to right: MRI images for four different modalities (Flair, T1ce, T1, and T2) along with their corresponding labels.}
    \label{Figure: 4_modality}
\end{figure}

Brain tumor segmentation is essential for providing valuable information such as tumor location and size for diagnosis and surgical planning. Clinically, doctors usually perform tumor segmentation~\cite{ding2021rfnet,ZHOU2022108417,chen2019robust} based on multi-modal MRI images with four modalities as shown in Fig.~\ref{Figure: 4_modality}, including T1-weighted (T1), T2-weighted (T2), contrast-enhanced T1-weighted (T1ce), and Fluid Attenuation Inversion Recovery (Flair). However, in practice, the common problem is that some modalities may not be available due to varying scanning protocols and patient conditions. Thus, segmentation from incomplete MRI modalities becomes a challenging issue.

Previous works~\cite{have2016,ALPAR2022108675} have attempted to address the challenge of incomplete multi-modal brain tumor segmentation. Havaei et al~\cite{have2016} and Dorent~\cite{dorent2019hetero} fused the mean and variance of accessible multi-modal features to overcome the missing modality issue. However, this fusion treats each modality equally without considering different missing scenarios, which may fail to aggregate features effectively. Chen~\etal~\cite{chen2019robust} and Zhou~\etal~\cite{ALPAR2022108675} leveraged attention mechanisms to extract information between modalities and fuse enhanced features using the same attention mechanisms. However, these methods do not fully exploit the relations between tumor regions and image modalities.

\begin{figure*}[ht]
  \centering
    \includegraphics[width=\linewidth]{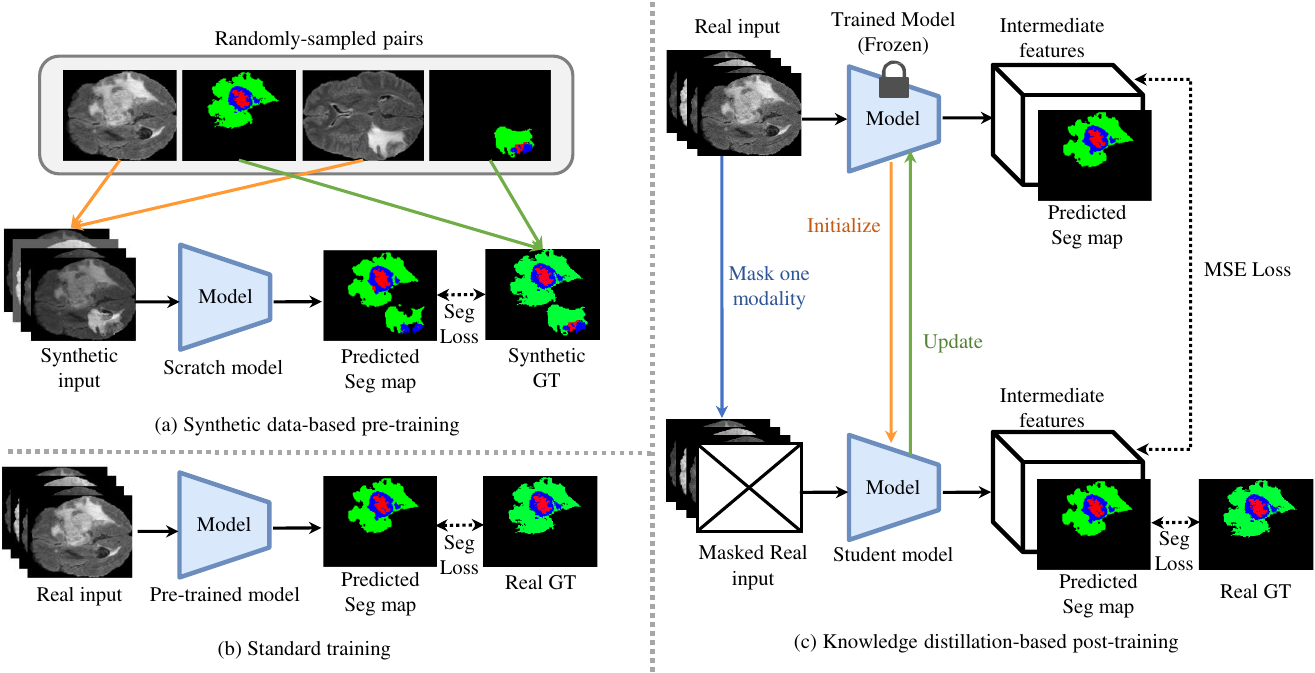}
    \caption{Overview of the proposed approach. `Seg' is short for `segmentation'. (a) Pre-train the model from scratch with the synthetic data by exploiting the asymmetry of brain MRI images. (b) Conduct standard training with real training samples based on the pre-trained model. (c) Conduct distillation-based post-training to improve the model's ability to deal with partial-modalitty inputs.}
    \label{Figure-knowledge}
\end{figure*}

To address this limitation, RFNet~\cite{ding2021rfnet} applied different attention to different modalities to improve brain tumor segmentation performance. 
Another strategy for incomplete multimodal learning is synthesizing the missing modalities using generative models~\cite{van2015does}. However, these methods are limited by the ability of the generative models.
Zhang et al.~\cite{zhang2021modality} proposed an ensemble learning of single-modal models with adaptive fusion to achieve multimodal segmentation. However, it only works when one or all modalities are available. Although these methods aim to recover the missing modality information, they have ignored the intrinsic problems of medical image segmentation:
one of the main challenges of BTS is the limited availability of training samples, particularly for cases with tumors. This makes manual annotation of images time-consuming and resource-intensive.
In general, image segmentation models can be trained with pre-trained models that were trained on object classification datasets, such as ImageNet ~\cite{imagenet}, to alleviate the problem of limited training samples. However, BTS models do not have a commonly used pre-training dataset due to the high cost of acquiring medical data. To address this challenge, we propose a pre-training stage that uses synthesized samples to improve the model's ability to adapt to different brain anatomies and tumor shapes.
As shown in Fig.~\ref{Figure:sample_synthesis}, we observed that the brain structure is often symmetrical except for regions with tumors. We leverage this asymmetry for model pre-training. We first calculate the brain image's symmetrical line and compare the left and right parts of the image. The area of difference is considered as the tumor parts and is appended to healthy brain images to form synthetic training sample images. This approach allows us to synthesize a pre-training dataset with a large diversity that covers a wide range of different combinations of tumor shapes and brain anatomy.

Alternatively, some methods explore knowledge distillation from complete modalities to incomplete ones~\cite{chen2021learning,hu2020knowledge,wang2021acn}. Although promising results have been obtained, such methods require training and deploying a specific model for each subset of missing modalities. 

To overcome this challenge, we propose a post-training stage to enable the BTS model to reconstruct the missing modality in the prediction results when only partial modalities are given. After a standard training step, the trained model is copied into a teacher model and a student model. Then, for the input modalities to the teacher model, we randomly remove one of the modalities to form the inputs to the student model. During the post-training stage, we fine-tune the student model to generate similar outputs as the teacher model, so that the student model can progressively learn to recover missing information in the prediction. This post-training stage can be viewed as a knowledge distillation (KD)~\cite{hinton2015distilling} process that trains the student model to recover missing modal information from the existing modal information acquired by the teacher model.

To improve brain tumor segmentation (BTS) performance, we propose a pre-training stage that utilizes synthetic data to decouple tumors from brain anatomy, and a post-training stage that utilizes knowledge distillation (KD) to recover missing modalities in the final prediction. Our main contributions are:
\begin{itemize}
\item We propose a synthetic data-based pre-training strategy to help the BTS model form prior knowledge for segmenting tumors out of various brain anatomy. This allows the BTS model to be easily trained and improved without manual label efforts.
\item We propose a KD-based post-training strategy that learns to recover missing modal information from existing modal information.
\item The performance on BRATS2020, BRATS2018, and BRATS2015 datasets demonstrate that our methods outperform the state-of-the-art and achieve new state-of-the-art.
\end{itemize}
\section{Related Work}
\subsection{Semantic segmentation}
Semantic segmentation is a fundamental task in computer vision that assigns a class label to each pixel in an image. With advancements in deep learning, current state-of-the-art methods~\cite{chen2018encoder,liu2020guided,liu2020crnet,fu2019dual,liu2021few} typically adopt fully convolutional networks~\cite{long2015fully} to make dense predictions. To improve feature representation, the encoder-decoder architecture~\cite{long2015fully,deeplab} has become popular.
In the encoder-decoder structure, the encoder gradually reduces the size of feature maps, acquiring a broad field of view and extracting abstract feature representations. The decoder then recovers fine-grained information. Skip connections are often utilized to integrate high-level and low-level features for improved predictions.
The Deeplab~\cite{deeplab} takes this approach one step further by incorporating an atrous spatial pyramid pooling scheme, which fuses features from different fields of view using dilated convolution networks to further increase the field of view and improve prediction results.
Our method also adopts the encoder-decoder architecture to transfer cross-modality information in low-resolution feature maps and utilizes decoders to recover details.

\subsection{Incomplete Multi-modal Tumor Segmentation}
In clinical settings, incomplete imaging data is a common occurrence, often caused by missing modalities~\cite{zhao2019data,wang2020lt,zhao2019data,zhao2022modality} and scarce annotations~\cite{tran2017missing,liu2021incomplete,zhi2020factorized}. This study focuses on incomplete modalities in Brain Tumor Segmentation (BTS). Unlike traditional tumor segmentation methods that use full modalities of MRI data, BTS from incomplete data is more challenging.
Previous studies have attempted to address this issue. Shen~\etal~\cite{ZHOU2022108417} used adversarial learning to project images from existing modalities into a unified feature space during segmentation. Zhou~\etal~\cite{ALPAR2022108675} generated missing features by utilizing correlations between different modalities. However, these methods are not suitable when multiple modalities are missing.
Several other approaches~\cite{dorent2019hetero,chen2019robust} were proposed to aggregate available modalities as fused features to address missing ones. However, these methods did not fully consider the relationships between brain tumor regions. To improve region-fusing efficiency, RFNet~\cite{ding2021rfnet} proposed fusing features at the region level.

In this study, we also aim to recover missing modalities from the existing ones using a KD-based post-training stage. We combine the recovered features with existing features using region-level fusing for improved results.

\subsection{Data Augmentation}
The limited size of datasets has always posed a challenge in medical imaging. To overcome this, data augmentation techniques such as random flipping, mirroring, and rotation have been widely adopted to obtain highly generalized deep models~\cite{liu2021cross,pspnet,ding2021rfnet,liu2024harmonizing}. Some recent studies~\cite{2018arXiv180509501C,zhong2017random} have identified better augmentation strategies from a pool of methods.
Mixup~\cite{mixup} has been shown to improve long-tailed classification tasks by diversifying tail classes with features from head classes in recent studies~\cite{bbn,zhong2021improving,zhang2020tricks}.
In this paper, we propose a new asymmetry-based approach for synthesizing training samples for the brain tumor segmentation (BTS) task, which can generate high-quality synthetic images and simplify the training of the models.

\begin{figure}[t]
  \centering
    \includegraphics[width=0.7\linewidth]{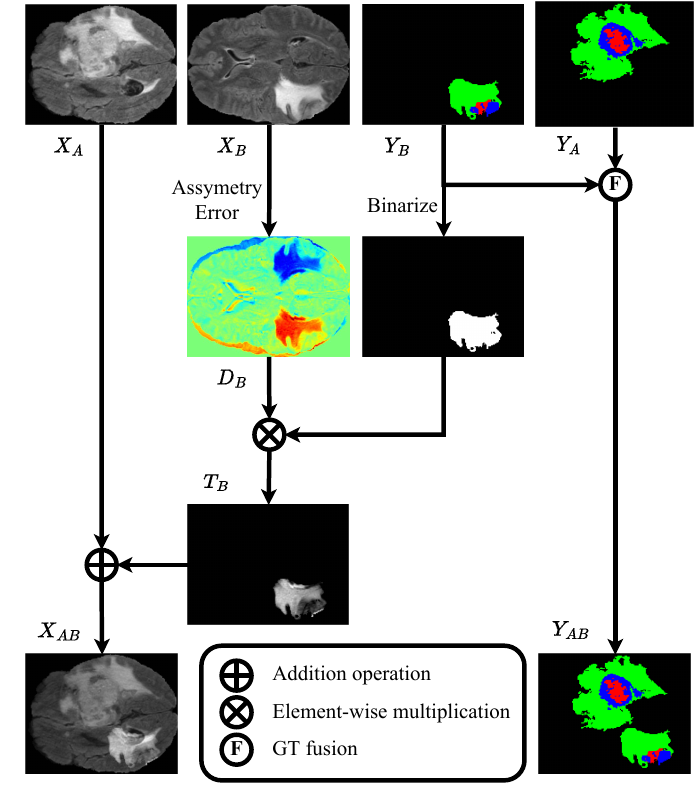}
    \caption{Proposed sample synthesis process for generating the pre-training dataset.}
    \label{Figure:sample_synthesis}
\end{figure}

\section{Method}
The proposed approach addresses two main challenges in previous multi-modal brain tumor segmentation (BTS) methods: 1) poor generalization to different brain anatomy and 2) difficulty in adapting to partial-modality inputs. In addition to the standard training stage (as shown in Fig. ~\ref{Figure-knowledge}(b)), our approach introduces an extra pre-training stage and post-training stage. To address challenge 1, we propose a pre-training stage using synthetic data (as shown in Fig. ~\ref{Figure-knowledge}(a)). To address challenge 2, we propose a post-training stage based on knowledge distillation (KD) to enable the model to learn how to recover missing information through KD-based fine-tuning (as shown in Fig. ~\ref{Figure-knowledge}(c)).

\subsection{Task Definition}
\begin{figure*}[htbp]
  \centering
    \includegraphics[width=1\linewidth]{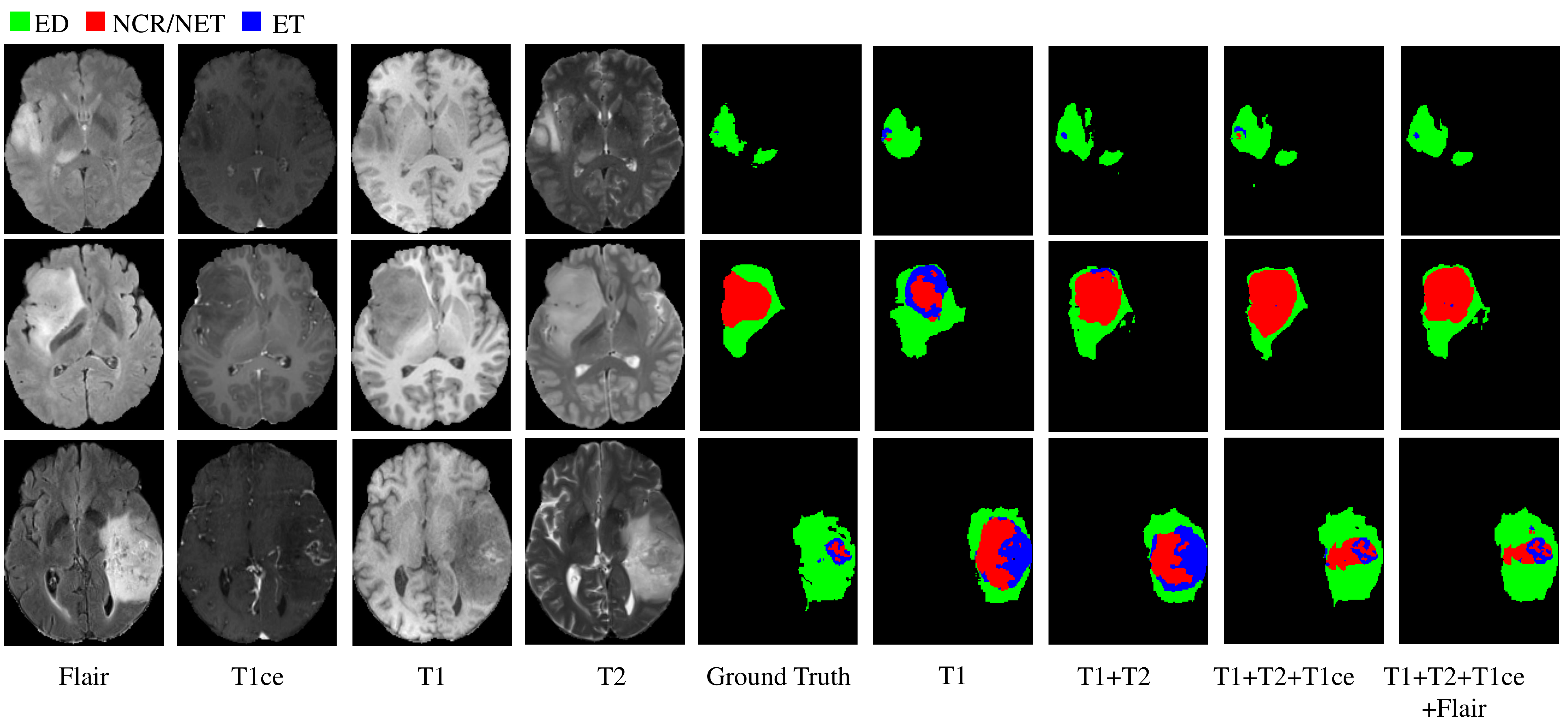}
    \caption{The segmentation results of our method using different MRI modalities. It can be observed that as the number of modalities increases, our method is able to predict more accurate masks.}
    \label{Figure: different_modal_prediction}
\end{figure*}

Incomplete multi-modal brain tumor segmentation involves identifying the regions of the tumor in multi-modal MRI images, including the whole tumor, tumor core, and enhancing tumor. Fig.~\ref{Figure: 4_modality} illustrates the four types of modal MRI: 1) Native MRI (T1), 2) Post-Contrast T1-Weighted MRI (T1ce), 3) T2-Weighted MRI (T2), and 4) T2 Fluid Attenuated Inversion Recovery (FLAIR). As depicted in Fig.~\ref{Figure: different_modal_prediction}, the whole tumor region must encompass all tumor regions, including necrotic and non-enhancing tumor core (NCR/NET), peritumoral edema (ED), and GD-enhancing tumor (ET). The tumor core region must encompass NCR/NET and ET, while the enhancing tumor region only covers the ET.

\subsection{Pre-training with Synthetic Data Based on Asymmetry of Brain MRI Images}
We propose a novel data synthesis method that leverages the asymmetry of brain tumor images to generate a pre-training dataset with Asymmetric Error Maps (AEM), preparing the BTS model to recognize tumor shapes across various brain anatomy. The complete process is depicted in Fig.~\ref{Figure:sample_synthesis}. To create one sample, we extract the tumor region from $X_B$ and embed it into the anatomy of $X_A$. It is important to note that $(X_A, Y_A)$ and $(X_B, Y_B)$ are two separate real samples, with $X_A, X_B$ being MRI images and $Y_A, Y_B$ being their respective ground truth annotations.

Since a tumor on an MRI image typically presents higher intensities than normal regions, an MRI image with a tumor region can be denoted as:
\begin{equation}
\label{eq:decouple}
    X_B = \Tilde{X}_B + T_B,
\end{equation}
where $\Tilde{X}_B $ is the healthy counterpart of $X_B$, and $T_B$ is the additional intensities brought by the tumor. 
Accordingly, the MRI image is decomposed into an anatomy part, $\Tilde{X}_B$, and a tumor part, $T_B$. However, in practice, we do not have access to brain MRI images when the patient is healthy, so we cannot obtain the true $\Tilde{X}_B$. To separate $T_B$ from $X_B$, we leverage the symmetry property of a healthy human brain, which is symmetric when healthy. By comparing the two sides of the unhealthy brain image, we can estimate the additional intensity brought by the tumor. Thus, $T_B$ can be obtained by:
\begin{equation}
\begin{aligned}
    D_B = |X_B - vflip(X_B)|, \\
    T_B = D_B \bigotimes binarize(Y_B),
\end{aligned}
\end{equation}
where $vflip(\cdot)$ applies vertical flip\footnote{We simply adopt $vflip(\cdot)$ since all images in BRATS2020, BRATS2018, and BRATS2015 we used are placed horizontally. For future cases whose directions are unknown, methods for detecting the axis of symmetry ~\cite{prasad2004finding} can be adopted.} to the input image, and $binarize(\cdot)$ applies binarization to the input mask. Note that in practice, we calibrate the flipped image to minimize the error between brain outline edges of $X_B$ and $vflip(X_B)$, while such error is rather low in the dataset we use. $D_B$ is the asymmetry error map to denote the asymmetry between the two sides of the brain caused by the tumor. Since $D_B$ presents an error on both sides, we further use the Ground Truth(GT) $Y_B$ to indicate the side of the tumor is grown. By combining $Y_B$ and $D_B$ with element-wise product ($\bigotimes$), $T_B$ can be obtained.
Then we add $T_B$ to another image $X_A$ to imitate the process by which a similar tumor is grown in another brain anatomy:
\begin{equation}
\label{eq:synthesize}
    X_{AB} = X_A + T_B,
\end{equation}
which denotes the tumor $T_B$ is grown on $X_A$ and results in an imaging $X_{AB}$

Besides synthesizing brain MRI images, we also need to generate corresponding GT segmentation maps. 
Thus, during the GT fusing, we directly overlay the target tumor area ($Y_B$) to the source image GT map ($Y_A$). Suppose the tumor annotations of the two GT maps overlap in the space after fusing the two GT samples. The one that has a higher ranking value is taken according to the ranking of $ET > NCR/NET > ED$ \footnote{$ET$ represents enhancing tumor (\textcolor{blue}{blue} tumor areas), $NCR/NET$ represents necrotic components and non-enhancing tumor (\textcolor{red}{red} tumor areas), and $ED$ represents peritumoral edema (\textcolor{green}{green} tumor areas).}.
The ranking depends on the following:
\begin{equation}
WT(\text{Whole Tumor}) = ED + NCR/NET + ET;
\end{equation}
\begin{equation}
TC(\text{Tumor Core}) = NCR/NET + ET; 
\end{equation}
\begin{equation}
ET(\text{Enhance Tumor}) = ET. 
\end{equation}
Through the synthesis process, the synthesized $X_{AB}$ and $Y_{AB}$ are considered new data to use.

\begin{figure}[t]
  \centering
    \includegraphics[width=1\linewidth]{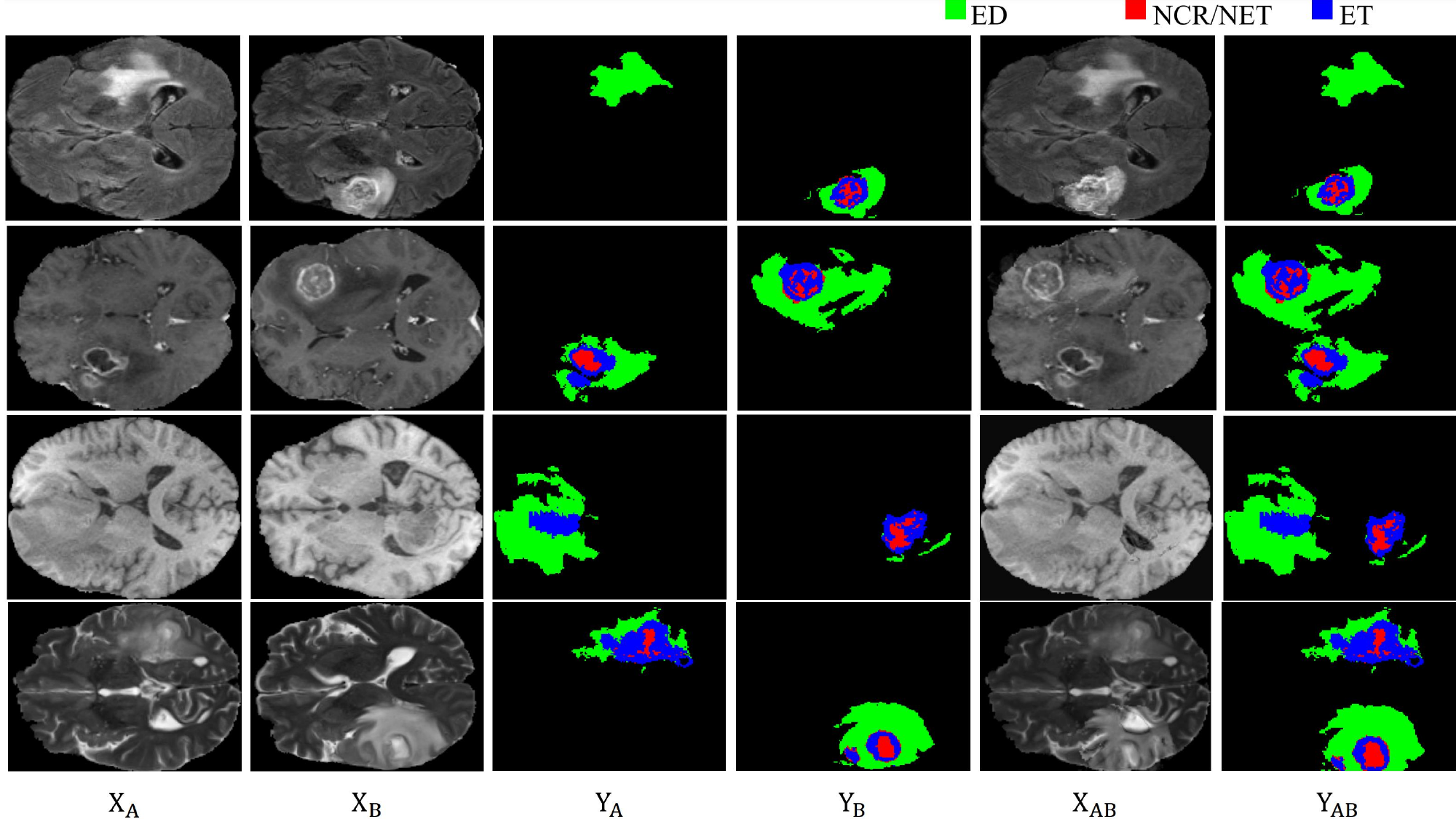}
    \caption{The synthetic samples. The tumor part of $X_B$ is embedded into $X_A$, which generates $X_{AB}$. The ground truth $Y_{AB}$ is generated by the fusion of $Y_A$ and $Y_B$. For each modality, we provide one example (Flair, T1ce, T1, and T2 are listed from top to bottom).}
    \label{Figure: Additional_synthetic_samples}
\end{figure}      

\begin{table*}[ht]
\Huge
\caption{Performances of our method and the state-of-the-art approaches on BRATS2020. The available and missing modalities are denoted by \textbf{$\bullet$} and \textbf{$\circ$}, respectively.}
\renewcommand\arraystretch{1} 
\begin{adjustbox}{width=\columnwidth,center}
\begin{tabular}{cccc|ccccc|ccccc|ccccc}
\toprule
\multicolumn{4}{c|}{}                                                                                                                                  & \multicolumn{15}{c}{\textbf{Dice score(\%)}}                                                                                                                                                                                                                                                                                                                                                                                                                                                                                                                                                                                         \\ \cline{5-19} 
\multicolumn{4}{c|}{\multirow{-2}{*}{\textbf{Modality}}}                                                                                               & \multicolumn{5}{c|}{\textbf{Whole}}                                                                                                                                                                        & \multicolumn{5}{c|}{\textbf{Core}}                                                                                                                                                                         & \multicolumn{5}{c}{\textbf{Enhancing}}                                                                                                                                                                     \\
\midrule

{\textbf{F}} & {\textbf{T1}} & {\textbf{T1ce}} & {\textbf{T2}} & {\textbf{HeMIS}} & {\textbf{UHVED}} & {\textbf{RobustSeg}} & {\textbf{RFNet}} & {\textbf{Ours}} & {\textbf{HeMIS}} & {\textbf{UHVED}} & {\textbf{RobustSeg}} & {\textbf{RFNet}} & {\textbf{Ours}} & {\textbf{HeMIS}} & {\textbf{UHVED}} & {\textbf{RobustSeg}} & {\textbf{RFNet}} & {\textbf{Ours}} \\
\midrule
\textbf{$\circ$}&\textbf{$\circ$}&\textbf{$\circ$}&\textbf{$\bullet$}
&79.85&80.85&81.29&86.18&\textbf{87.56}
&54.22&54.96&60.97&71.14&\textbf{73.48}
&31.43&32.05&38.25&47.53&\textbf{51.48}\\
\textbf{$\circ$}&\textbf{$\circ$}&\textbf{$\bullet$}&\textbf{$\circ$}
&64.58&66.48&67.31&77.68&\textbf{81.45}
&69.41&71.19&73.26&82.90&\textbf{84.14}
&63.24&60.48&65.64&77.01&\textbf{79.32}\\
\textbf{$\circ$}&\textbf{$\bullet$}&\textbf{$\circ$}&\textbf{$\circ$}
&63.01&59.56&60.25&76.00&\textbf{81.35}
&42.42&44.63&43.74&62.61&\textbf{69.95}
&16.53&11.90&21.23&37.70&\textbf{42.29}\\
\textbf{$\bullet$}&\textbf{$\circ$}&\textbf{$\circ$}&\textbf{$\circ$}
&52.29&77.18&82.65&86.79&\textbf{88.56}
&24.97&52.02&55.94&69.42&\textbf{71.91}
& 9.00&30.32&31.07&43.36&\textbf{44.23}\\
\textbf{$\circ$}&\textbf{$\circ$}&\textbf{$\bullet$}&\textbf{$\bullet$}
&84.45&82.30&85.00&88.19&\textbf{88.95}
&77.60&75.35&81.88&\textbf{85.82}&85.19
&70.30&65.54&72.08&78.16&\textbf{81.83}\\
\textbf{$\circ$}&\textbf{$\bullet$}&\textbf{$\bullet$}&\textbf{$\circ$}
&72.50&71.65&73.16&81.02&\textbf{83.71}
&75.59&76.35&79.47&83.89&\textbf{84.58}
&70.71&65.00&73.01&77.79&\textbf{81.45}\\
\textbf{$\bullet$}&\textbf{$\bullet$}&\textbf{$\circ$}&\textbf{$\circ$}
&65.29&82.56&86.54&89.15&\textbf{90.49}
&41.58&60.50&63.48&72.41&\textbf{74.14}
&13.99&32.83&37.35&47.22&\textbf{49.31}\\
\textbf{$\circ$}&\textbf{$\bullet$}&\textbf{$\circ$}&\textbf{$\bullet$}
&82.31&81.64&84.78&87.73&\textbf{88.83}
&56.38&60.77&65.41&73.19&\textbf{74.04}
&28.58&32.55&40.46&50.81&\textbf{53.92}\\
\textbf{$\bullet$}&\textbf{$\circ$}&\textbf{$\circ$}&\textbf{$\bullet$}
&81.56&82.19&87.27&89.26&\textbf{90.37}
&55.89&59.28&66.30&74.33&\textbf{75.83}
&28.91&35.62&42.77&53.67&\textbf{54.26}\\
\textbf{$\bullet$}&\textbf{$\circ$}&\textbf{$\bullet$}&\textbf{$\circ$}
&69.37&82.23&86.60&89.67&\textbf{90.42}
&70.86&73.61&80.80&85.50&\textbf{86.21}
&68.31&63.07&70.54&76.77&\textbf{80.44}\\
\textbf{$\bullet$}&\textbf{$\bullet$}&\textbf{$\bullet$}&\textbf{$\circ$}
&73.31&84.42&87.51&90.24&\textbf{91.05}
&75.07&76.97&82.53&85.65&\textbf{85.92}
&70.80&64.75&73.02&80.00&\textbf{82.01}\\
\textbf{$\bullet$}&\textbf{$\bullet$}&\textbf{$\circ$}&\textbf{$\bullet$}
&83.03&84.56&88.69&90.26&\textbf{90.87}
&57.40&63.38&68.33&74.88&\textbf{75.51}
&29.53&37.07&44.42&52.38&\textbf{52.79}\\
\textbf{$\bullet$}&\textbf{$\circ$}&\textbf{$\bullet$}&\textbf{$\bullet$}
&84.64&84.72&88.49&90.40&\textbf{91.15}
&77.69&75.62&82.20&85.88&\textbf{86.11}
&71.36&63.75&71.41&79.23&\textbf{82.41}\\
\textbf{$\circ$}&\textbf{$\bullet$}&\textbf{$\bullet$}&\textbf{$\bullet$}
&85.19&84.04&85.81&88.72&\textbf{89.33}
&79.05&77.68&82.86&\textbf{86.06}&84.92
&71.67&66.31&73.96&79.31&\textbf{82.30}\\
\textbf{$\bullet$}&\textbf{$\bullet$}&\textbf{$\bullet$}&\textbf{$\bullet$}
&85.19&85.84&89.03&90.95&\textbf{91.29}
&78.58&77.67&83.16&\textbf{86.00}&85.91
&71.49&64.92&73.03&80.66&\textbf{82.09}\\
\midrule
\multicolumn{4}{c|}{\textbf{Average}}
&75.10&79.06&82.29&86.81&\textbf{88.36}
&65.45&66.67&71.36&78.65&\textbf{79.86}
&47.73&48.41&55.22&64.11&\textbf{66.67}\\
\bottomrule
\end{tabular}
\end{adjustbox}

\label{table:soa_2020}
\end{table*}  
\begin{figure*}[ht]
  \centering
    \includegraphics[width=0.99\linewidth]{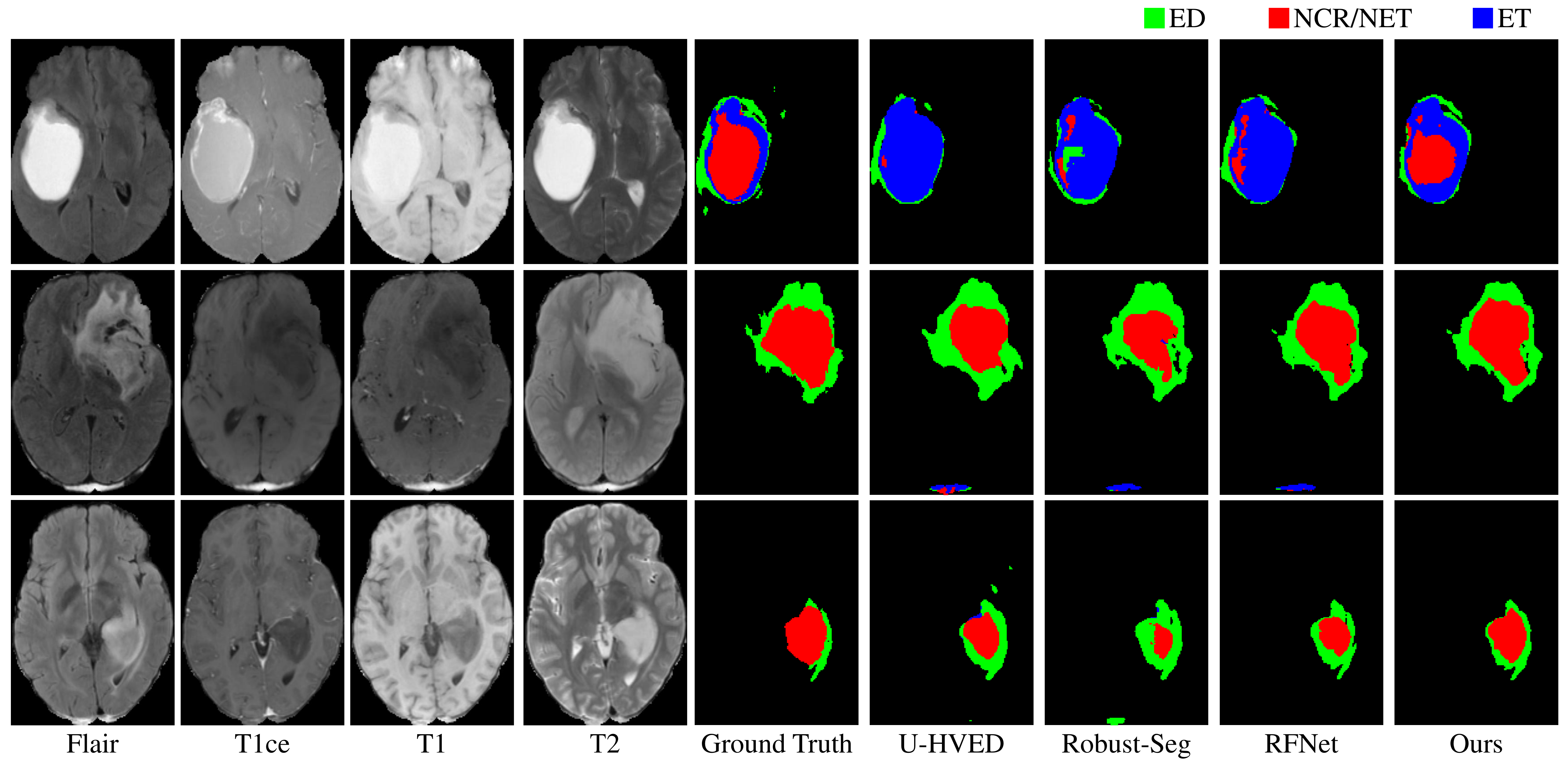}
    \caption{The visualization of segmentation results on BRATS2020 between our method and state-of-the-art methods. From left to right, we demonstrate the MRI images for the Flair modality, T1ce modality, T1 modality, and T2 modality. Following the ground truth, the prediction of U-HVED\cite{dorent2019hetero}, RobustSeg\cite{chen2019robust}, RFNet\cite{ding2021rfnet} and our prediction. 
    }
    \label{Figure: different_method_prediction}
\end{figure*}     

\begin{table*}[h]
\caption{Performances of our method and the state-of-the-art approaches on BRATS2018. The available and missing modalities are denoted by \textbf{$\bullet$} and \textbf{$\circ$}, respectively. }
\Huge
\renewcommand\arraystretch{1} 
\begin{adjustbox}{width=\columnwidth,center}
\begin{tabular}{cccc|ccccc|ccccc|ccccc}
\toprule
\multicolumn{4}{c|}{}                                                                                                                                  & \multicolumn{15}{c}{\textbf{Dice score(\%)}}                                                                                                                                                                                                                                                                                                                                                                                                                                                                                                                                                                                         \\ \cline{5-19} 
\multicolumn{4}{c|}{\multirow{-2}{*}{\textbf{Modality}}}                                                                                               & \multicolumn{5}{c|}{\textbf{Whole}}                                                                                                                                                                        & \multicolumn{5}{c|}{\textbf{Core}}                                                                                                                                                                         & \multicolumn{5}{c}{\textbf{Enhancing}}                                                                                                                                                                     \\
\midrule
{\textbf{F}} & {\textbf{T1}} & {\textbf{T1ce}} & {\textbf{T2}} & {\textbf{HeMIS}} & {\textbf{UHVED}} & {\textbf{RobustSeg}} & {\textbf{RFNet}} & {\textbf{Ours}} & {\textbf{HeMIS}} & {\textbf{UHVED}} & {\textbf{RobustSeg}} & {\textbf{RFNet}} & {\textbf{Ours}} & {\textbf{HeMIS}} & {\textbf{UHVED}} & {\textbf{RobustSeg}} & {\textbf{RFNet}} & {\textbf{Ours}} \\
\midrule
\textbf{$\circ$}     & \textbf{$\circ$}  & \textbf{$\circ$}    & \textbf{$\bullet$}  
&79.20 &80.90 &82.24 &84.30  &\textbf{85.89} &40.18 &54.10 &57.49 &67.62 &\textbf{72.76} &20.31 &30.80 &28.97 &40.71&\textbf{48.94}\\
\textbf{$\circ$}     & \textbf{$\circ$}  & \textbf{$\bullet$}    & \textbf{$\circ$}  
&58.50 &62.40 &73.31 &74.93 &\textbf{77.77} &44.55 &66.70 &76.83 &80.99 &\textbf{82.23}&49.93 &65.50 &67.07 &69.43& \textbf{75.63}\\

\textbf{$\circ$}     & \textbf{$\bullet$}  & \textbf{$\circ$}    & \textbf{$\circ$}  
&54.30 &52.40 &70.11 &74.68& \textbf{77.40} &17.42 &37.20 &47.90 &64.42 &\textbf{66.51}&4.67 &13.70 &17.29 &34.43&\textbf{37.43}\\

\textbf{$\bullet$}     & \textbf{$\circ$}  & \textbf{$\circ$}    & \textbf{$\circ$}  
&79.90 &82.10 &85.69 &86.46 &\textbf{88.43}&37.45 &50.40 &53.57 &64.89 &\textbf{71.80}&5.57 &24.80 &25.69 &33.92 &\textbf{40.94}\\

\textbf{$\circ$}     & \textbf{$\circ$}  & \textbf{$\bullet$}    & \textbf{$\bullet$} 
&81.00 &82.70 &85.19 &86.39  &\textbf{86.68}&63.39 &73.70 &80.20 &83.27&\textbf{84.84} &65.38 &70.20 &69.71 &73.01& \textbf{77.68}\\
\textbf{$\circ$}     & \textbf{$\bullet$}  & \textbf{$\bullet$}    & \textbf{$\circ$} 
&63.80 &66.80 &77.18 &78.59  &\textbf{80.27}&55.06 &69.70 &78.72 &82.22& \textbf{82.74} &62.40 &67.00 &69.06 &70.73& \textbf{72.71}\\

\textbf{$\bullet$}     & \textbf{$\bullet$}  & \textbf{$\circ$}    & \textbf{$\circ$}  
 &83.90 &84.30 &88.24 &88.78& \textbf{89.18} &49.52 &55.30 &60.68 &71.59 &\textbf{74.83}&22.26 &24.20 &32.13 &39.68& \textbf{45.31}\\

\textbf{$\circ$}     & \textbf{$\bullet$}  & \textbf{$\circ$}    & \textbf{$\bullet$}  
&80.80 &82.20 &84.78 &86.15 &\textbf{86.89}&47.26 &57.20 &62.19 &70.89  &\textbf{74.26}&23.56 &30.70 &32.01 &41.42& \textbf{48.17}\\

\textbf{$\bullet$}     & \textbf{$\circ$}  & \textbf{$\circ$}    & \textbf{$\bullet$}  
&86.00 &87.50 &88.28 &89.12 &\textbf{89.65}&53.42 &59.70 &61.16 &70.82& \textbf{74.70}  &23.19 &34.60 &33.84 &43.77& \textbf{50.78}\\

\textbf{$\bullet$}     & \textbf{$\circ$}  & \textbf{$\bullet$}    & \textbf{$\circ$} 
&83.30 &85.50 &88.51 &89.17  &\textbf{89.76}&66.12 &72.90 &80.62 &82.94 &\textbf{85.07} &67.12 &70.30 &70.30 &72.84& \textbf{74.12} \\

\textbf{$\bullet$}     & \textbf{$\bullet$}  & \textbf{$\bullet$}    & \textbf{$\circ$}  
&85.10 &86.20 &88.73 &89.71& \textbf{89.77}  &69.26 &74.20 &81.06 &83.77& \textbf{84.79} &71.30 &71.10 &70.78 &73.17& \textbf{74.15}\\

\textbf{$\bullet$}     & \textbf{$\bullet$}  & \textbf{$\circ$}    & \textbf{$\bullet$}  
&87.00 &88.00 &88.81 &89.68 &\textbf{89.82}&57.76 &61.50 &64.38 &73.09 &\textbf{76.06}&28.46 &34.10 &36.41 &44.79 &\textbf{50.66} \\

\textbf{$\bullet$}     & \textbf{$\circ$}  & \textbf{$\bullet$}    & \textbf{$\bullet$} 
&87.00 &88.60 &89.27 &90.06& \textbf{90.31} &70.62 &75.60 &80.72 &83.54 &\textbf{84.75}&70.52 &71.20 &70.88 &73.13& \textbf{76.17}\\

\textbf{$\circ$}     & \textbf{$\bullet$}  & \textbf{$\bullet$}    & \textbf{$\bullet$}  
&82.10 &83.30 &86.01 &86.78& \textbf{86.83} &66.60 &75.30 &80.33 &83.97 &\textbf{84.70}&67.84 &71.10 &70.10 &72.56& \textbf{76.24}\\

\textbf{$\bullet$}     & \textbf{$\bullet$}  & \textbf{$\bullet$}    & \textbf{$\bullet$}  
&87.60 &88.80 &89.45 &90.26& \textbf{90.30} &72.50 &76.40 &80.86 &84.02 & \textbf{84.84}&75.37 &71.70 &71.13 &73.21& \textbf{76.08}\\
\midrule
\multicolumn{4}{c|}{\textbf{Average}}                       
&78.60 &80.10 &84.39 &85.67& \textbf{86.60} &54.07 &64.00 &69.78 &76.53& \textbf{78.99} &43.86 &50.00 &51.02 &57.12& \textbf{61.67}\\

\bottomrule
\end{tabular}
\end{adjustbox}

\label{table:soa_2018}
\end{table*}               
\begin{table*}[h]
\caption{Performances of our method and the state-of-the-art approaches on BRATS2015. The available and missing modalities are denoted by \textbf{$\bullet$} and \textbf{$\circ$}, respectively.}
\Huge
\renewcommand\arraystretch{1} 
\begin{adjustbox}{width=\columnwidth,center}
\begin{tabular}{cccc|ccccc|ccccc|ccccc}
\toprule
\multicolumn{4}{c|}{}                                                                                                                                  & \multicolumn{15}{c}{\textbf{Dice score(\%)}}                                                                                                                                                                                                                                                                                                                                                                                                                                                                                                                                                                                         \\ \cline{5-19} 
\multicolumn{4}{c|}{\multirow{-2}{*}{\textbf{Modality}}}                                                                                               & \multicolumn{5}{c|}{\textbf{Whole}}                                                                                                                                                                        & \multicolumn{5}{c|}{\textbf{Core}}                                                                                                                                                                         & \multicolumn{5}{c}{\textbf{Enhancing}}                                                                                                                                                                     \\
\midrule
 
\textbf{F} & \textbf{T1} & \textbf{T1ce} & \textbf{T2} & {  \textbf{HeMIS}} & \textbf{UHVED} & \textbf{RobustSeg} & \textbf{RFNet} & \textbf{Ours} & {  \textbf{HeMIS}} & \textbf{UHVED} & \textbf{RobustSeg} & \textbf{RFNet} & \textbf{Ours} & {  \textbf{HeMIS}} & \textbf{UHVED} & \textbf{RobustSeg} & \textbf{RFNet} & \textbf{Ours} \\
\midrule
  
\textbf{$\circ$} & \textbf{$\circ$} & \textbf{$\circ$} & \textbf{$\bullet$} 
&58.48 &81.19 &85.49 &86.89 &\textbf{87.72} &40.18 &53.40 &58.66 &63.81 &\textbf{68.79} &20.31 &29.05 &37.66 &40.07 &\textbf{44.50}\\

\textbf{$\circ$}     & \textbf{$\circ$}  & \textbf{$\bullet$}    & \textbf{$\circ$}
&33.46 &67.48 &71.86 &74.95 &\textbf{77.05} &44.55 &68.24 &72.87 &72.64 &\textbf{74.93} &49.93 &71.54 &70.22 &\bf{81.40} &79.48\\

\textbf{$\circ$} & \textbf{$\bullet$}  & \textbf{$\circ$} & \textbf{$\circ$} 
&33.22 &53.58 &68.40 &74.20 &\textbf{77.40} &17.42 &41.14 &50.00 &61.27 &\textbf{64.38} &4.67 &19.16 &22.67 &29.44 &\textbf{36.65}\\

\textbf{$\bullet$}     & \textbf{$\circ$}  & \textbf{$\circ$}    & \textbf{$\circ$}
&71.26 &83.82 &83.02 &86.91 &\textbf{88.59} &37.45 &51.37 &46.67 &58.71 &\textbf{62.62} &5.57 &22.18 &28.30 &35.23 &\textbf{42.06}\\

\textbf{$\circ$}     & \textbf{$\circ$}  & \textbf{$\bullet$}    & \textbf{$\bullet$} 
&67.59 &84.77 &87.53 &88.39 &\textbf{88.97} &63.39 &73.18 &78.46 &77.50 &\textbf{78.98} &65.38 &83.54 &76.82 &86.97 &\textbf{86.42}\\

\textbf{$\circ$}     & \textbf{$\bullet$}  & \textbf{$\bullet$}    & \textbf{$\circ$} 
&45.93 &69.65 &74.59 &78.13 &\textbf{79.46} &55.06 &68.85 &76.40 &74.06 &\textbf{76.10} &62.40 &76.96 &73.95 &\textbf{82.48} &82.46\\
    
\textbf{$\bullet$}     & \textbf{$\bullet$}  & \textbf{$\circ$}    & \textbf{$\circ$} 
&80.28 &85.82 &87.66 &88.51 &\textbf{89.20} &49.52 &58.39 &60.17 &66.88 &\textbf{69.36} &22.26 &26.65 &35.28 &40.95 &\textbf{46.81}\\

\textbf{$\circ$}     & \textbf{$\bullet$}  & \textbf{$\circ$}    & \textbf{$\bullet$} 
&69.56 &82.17 &87.87 &88.25 &\textbf{89.12} &47.26 &57.58 &64.88 &67.24 &\textbf{70.06} &23.56 &33.94 &41.05 &40.58 &\textbf{45.34}\\

\textbf{$\bullet$}     & \textbf{$\circ$}  & \textbf{$\circ$}    & \textbf{$\bullet$}  
&82.10 &87.74 &89.08 &89.62 &\textbf{90.29} &53.42 &59.13 &63.51 &68.74 &\textbf{68.28} &23.19 &30.31 &39.72 &44.64 &\textbf{44.54}\\

\textbf{$\bullet$}     & \textbf{$\circ$}  & \textbf{$\bullet$}    & \textbf{$\circ$} 
&79.80 &87.48 &88.01 &88.45 &\textbf{88.86} &66.12 &74.27 &78.09 &79.30 &\textbf{79.91} &67.12 &84.30 &76.62 &86.15 &\textbf{86.24}\\

\textbf{$\bullet$}     & \textbf{$\bullet$}  & \textbf{$\bullet$}    & \textbf{$\circ$}  
&80.88 &87.91 &87.73 &88.75 &\textbf{88.92} &69.26 &75.82 &80.68 &80.46 &\textbf{81.30} &71.30 &84.33 &78.81 &87.30 &\textbf{87.34}\\

\textbf{$\bullet$}     & \textbf{$\bullet$}  & \textbf{$\circ$}    & \textbf{$\bullet$}
&83.87 &87.59 &89.07 &89.93 &\textbf{90.75} &57.76 &62.43 &65.99 &69.75 &\textbf{70.32} &28.46 &33.21 &43.04 &44.21 &\textbf{46.63}\\

\textbf{$\bullet$}     & \textbf{$\circ$}  & \textbf{$\bullet$}    & \textbf{$\bullet$}  
&82.78 &89.85 &89.06 &90.07 &\textbf{90.82} &70.62 &75.10 &79.47 &79.29 &\textbf{80.30} &70.52 &86.03 &78.07 &\textbf{87.34} &87.25\\

\textbf{$\circ$}     & \textbf{$\bullet$}  & \textbf{$\bullet$}    & \textbf{$\bullet$}  
&70.98 &84.72 &88.26 &88.41 &\bf{89.27} &66.60 &74.85 &80.84 &79.18 &\bf{80.16} &67.84 &84.03 &78.56 &\textbf{87.47} &87.18\\

\textbf{$\bullet$}     & \textbf{$\bullet$}  & \textbf{$\bullet$}    & \textbf{$\bullet$}  
&83.15 &89.79 &89.07 &90.49 &\bf{90.72} &72.50 &76.48 &81.19 &80.16 &\bf{80.80} &75.37 &86.12 &79.13 &87.68 &\textbf{87.65}\\ \hline

\multicolumn{4}{c|}{\textbf{Average}} 
&68.22 &81.57 &84.45 &86.13 &\bf{86.97} &54.07 &64.68 &69.19 &71.93 &\bf{73.75} &43.86 &56.76 &57.33 &64.13 &\bf{65.86}\\

\bottomrule
\end{tabular}
\end{adjustbox}

\label{table:soa_2015}
\end{table*}         
\begin{table}[ht]
\caption{The effectiveness of each component of our method with different baselines on BRATS2020. The AEM denotes the pre-training with the proposed Asymmetric Error Maps, while the KD denotes Knowledge Distillation during post-training. All the experimental results are implemented using the author's code.}
\centering
\resizebox{0.6\textwidth}{!}{

\begin{tabular}{l|l|l|l} 
\hline
\textbf{Method}        & \textbf{Whole}        & \textbf{Core}         & \textbf{Enhancing}     \\ 
\hline
UHVED        & 79.06        & 66.67        & 48.41         \\
UHVED+AEM    & 79.61$_{\color{red}{+0.55}}$ & 70.58$_{\color{red}{+3.91}}$ & 51.80$_{\color{red}{+3.39}}$   \\
UHVED+AEM+KD & 80.70$_{\color{red}{+1.64}}$  & 71.27$_{\color{red}{+4.60}}$  & 52.87$_{\color{red}{+4.46}}$  \\ 
\hline
RobustSeg        & 82.29        & 71.36        & 55.22         \\
RobustSeg+AEM    & 82.90$_{\color{red}{+0.61}}$  & 73.01$_{\color{red}{+1.65}}$ & 55.91$_{\color{red}{+0.69}}$  \\
RobustSeg+AEM+KD & 83.63$_{\color{red}{+1.34}}$ & 74.28$_{\color{red}{+2.94}}$ & 56.70$_{\color{red}{+1.48}}$   \\ 
\hline
RFNet         & 86.81         & 78.65        & 64.11         \\
RFNet+AEM     & 87.67$_{\color{red}{+0.86}}$ & 78.80$_{\color{red}{+0.15}}$  & 65.64$_{\color{red}{+1.53}}$  \\
RFNet+AEM+KD  & 88.36$_{\color{red}{+1.55}}$ & 79.86$_{\color{red}{+1.21}}$ & 66.67$_{\color{red}{+2.56}}$  \\
\hline
\end{tabular}

}

\label{table:method_with_ours}
\end{table}
\begin{figure*}[ht]
  \centering
    \includegraphics[width=\linewidth]{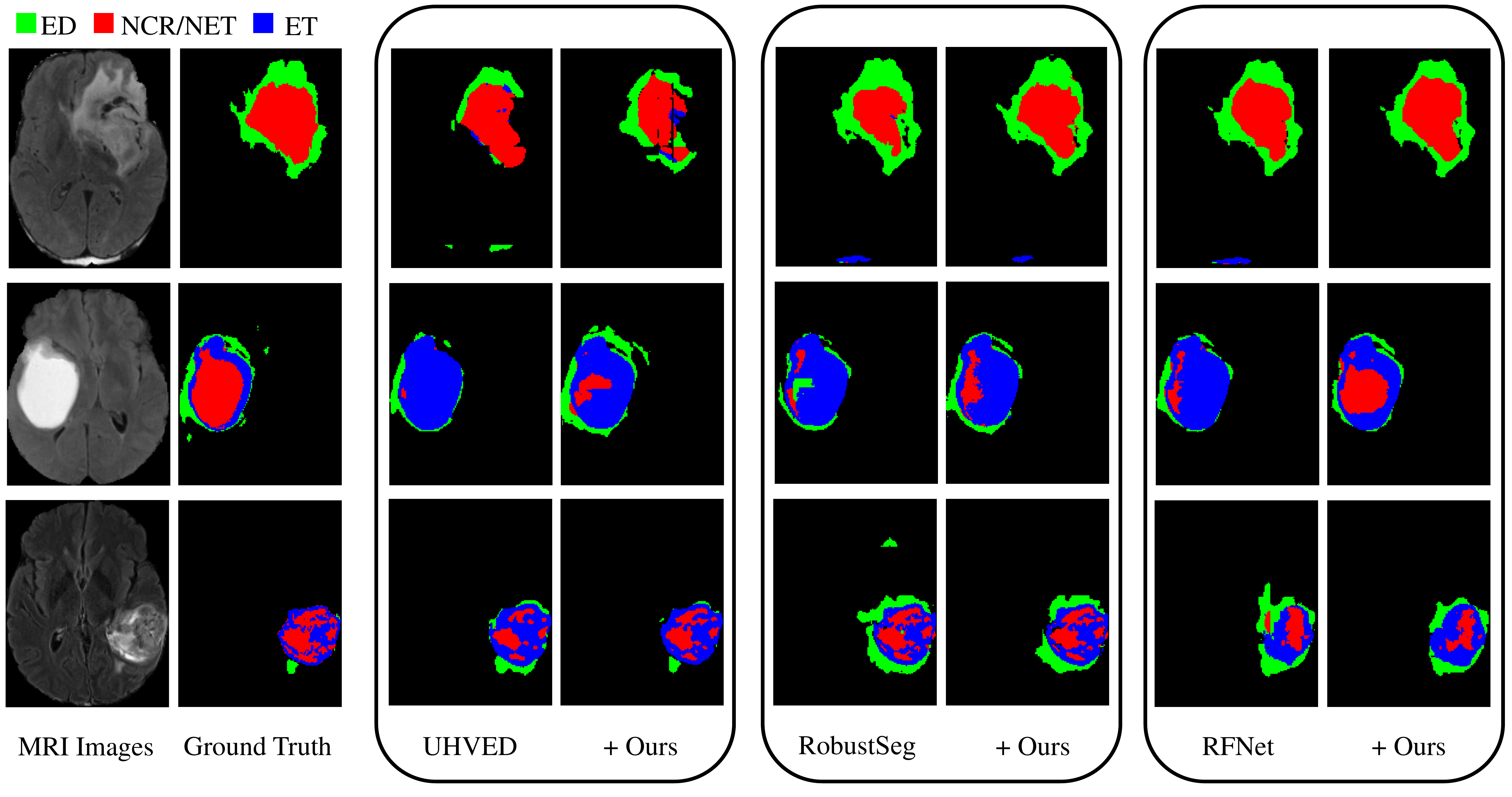}
    \caption{The visualization of segmentation results by adding our method to different baselines. Only the Flair modality is presented for MRI image visualization.}
    \label{Figure: methods_with_ours}
\end{figure*}     
\begin{table*}[h]
\caption{Detailed comparison of the performance of our proposed method with different baselines on the BRATS2020 dataset. The methods of UHVED\cite{dorent2019hetero}, RobustSeg\cite{chen2019robust}, and RFNet\cite{ding2021rfnet} are denoted by the abbreviations UH, RS, and RF, respectively. All experimental results were obtained using the code provided by the authors. And the available and missing modalities are denoted by \textbf{$\bullet$} and \textbf{$\circ$}, respectively.}
\Huge
\renewcommand\arraystretch{1} 
\begin{adjustbox}{width=\columnwidth,center}
\begin{tabular}{cccc|cccccc|cccccc|cccccc}
\toprule
\multicolumn{4}{c|}{}
&\multicolumn{18}{c}{\textbf{Dice score(\%)}}\\ 
\cline{5-22} 
\multicolumn{4}{c|}{\multirow{-2}{*}{\textbf{Modality}}} 
&\multicolumn{6}{c|}{\textbf{Whole}}
& \multicolumn{6}{c|}{\textbf{Core}}                                       & \multicolumn{6}{c}{\textbf{Enhancing}}\\
\midrule
{\textbf{F}} & {\textbf{T1}} & {\textbf{T1ce}} & {\textbf{T2}} & {\textbf{UH}} & {\textbf{UH+Ours}} & {\textbf{RS}} & {\textbf{RS+Ours}} & {\textbf{RF}} & {\textbf{RF+Ours}} & {\textbf{UH}} & {\textbf{UH+Ours}} & {\textbf{RS}} & {\textbf{RS+Ours}} & {\textbf{RF}} & {\textbf{RF+Ours}} & {\textbf{UH}} & {\textbf{UH+Ours}} & {\textbf{RS}} & {\textbf{RS+Ours}} & {\textbf{RF}} & {\textbf{RF+Ours}} \\
\midrule
\textbf{$\circ$}&\textbf{$\circ$}&\textbf{$\circ$}&\textbf{$\bullet$}
&76.57&80.85&81.29&82.39&86.18&\textbf{87.56}
&54.96&63.78&60.97&62.75&71.14&\textbf{73.48}
&32.05&38.88&38.25&38.85&47.53&\textbf{51.48}\\
\textbf{$\circ$}&\textbf{$\circ$}&\textbf{$\bullet$}&\textbf{$\circ$}
&66.48&67.41&67.31&69.91&77.68&\textbf{81.45}
&71.19&75.15&73.26&77.96&82.90&\textbf{84.14}
&60.48&65.31&65.64&69.42&77.01&\textbf{79.32}\\
\textbf{$\circ$}&\textbf{$\bullet$}&\textbf{$\circ$}&\textbf{$\circ$}
&59.56&61.69&60.25&68.06&76.00&\textbf{81.35}
&44.63&50.50&43.74&53.74&62.61&\textbf{69.95}
&11.90&24.24&21.23&26.43&37.70&\textbf{42.29}\\
\textbf{$\bullet$}&\textbf{$\circ$}&\textbf{$\circ$}&\textbf{$\circ$}
&77.18&80.74&82.65&80.27&86.79&\textbf{88.56}
&52.02&57.91&55.94&56.87&69.42&\textbf{71.91}
&30.32&30.98&31.07&32.26&43.36&\textbf{44.23}\\
\textbf{$\circ$}&\textbf{$\circ$}&\textbf{$\bullet$}&\textbf{$\bullet$}
&82.30&83.32&85.00&86.29&88.19&\textbf{88.95}
&75.35&80.68&81.88&84.27&\textbf{85.82}&85.19
&65.54&68.99&72.08&74.77&78.16&\textbf{81.83}\\
\textbf{$\circ$}&\textbf{$\bullet$}&\textbf{$\bullet$}&\textbf{$\circ$}
&71.65&70.93&73.16&75.25&81.02&\textbf{83.71}
&76.37&77.82&79.47&80.39&83.89&\textbf{84.58}
&65.00&67.74&73.01&74.41&77.79&\textbf{81.45}\\
\textbf{$\bullet$}&\textbf{$\bullet$}&\textbf{$\circ$}&\textbf{$\circ$}
&82.56&83.80&86.54&87.07&89.15&\textbf{90.49}
&60.50&64.05&63.48&67.87&72.41&\textbf{74.14}
&32.83&34.94&37.35&38.79&47.22&\textbf{49.31}\\
\textbf{$\circ$}&\textbf{$\bullet$}&\textbf{$\circ$}&\textbf{$\bullet$}
&81.64&82.08&84.78&86.11&87.73&\textbf{88.83}
&60.77&65.27&65.41&68.69&73.19&\textbf{74.04}
&32.55&38.69&40.46&42.50&50.81&\textbf{53.92}\\
\textbf{$\bullet$}&\textbf{$\circ$}&\textbf{$\circ$}&\textbf{$\bullet$}
&82.19&84.99&87.27&87.53&89.26&\textbf{90.37}
&59.28&65.85&66.30&68.31&74.33&\textbf{75.83}
&35.62&41.21&42.77&42.33&53.67&\textbf{54.26}\\
\textbf{$\bullet$}&\textbf{$\circ$}&\textbf{$\bullet$}&\textbf{$\circ$}
&82.23&84.86&86.60&87.42&89.67&\textbf{90.42}
&73.61&78.42&80.80&83.30&85.50&\textbf{86.21}
&63.07&67.21&70.54&70.76&76.77&\textbf{80.44}\\
\textbf{$\bullet$}&\textbf{$\bullet$}&\textbf{$\bullet$}&\textbf{$\circ$}
&84.42&85.69&87.51&88.30&90.24&\textbf{91.05}
&76.97&79.84&82.53&84.10&85.65&\textbf{85.92}
&64.75&68.34&73.02&72.80&80.00&\textbf{82.01}\\
\textbf{$\bullet$}&\textbf{$\bullet$}&\textbf{$\circ$}&\textbf{$\bullet$}
&84.56&86.24&88.69&89.36&90.26&\textbf{90.87}
&63.38&68.51&68.33&71.60&74.88&\textbf{75.51}
&37.07&41.27&44.42&44.27&52.38&\textbf{52.79}\\
\textbf{$\bullet$}&\textbf{$\circ$}&\textbf{$\bullet$}&\textbf{$\bullet$}
&84.72&86.91&88.49&89.55&90.40&\textbf{91.15}
&75.62&79.94&82.20&84.43&85.88&\textbf{86.11}
&63.75&67.78&71.41&73.27&79.23&\textbf{82.41}\\
\textbf{$\circ$}&\textbf{$\bullet$}&\textbf{$\bullet$}&\textbf{$\bullet$}
&84.04&83.61&85.81&87.07&88.72&\textbf{89.33}
&77.68&80.58&82.86&84.88&\textbf{86.06}&84.92
&66.31&69.19&73.96&75.23&79.31&\textbf{82.30}\\
\textbf{$\bullet$}&\textbf{$\bullet$}&\textbf{$\bullet$}&\textbf{$\bullet$}
&85.84&87.33&89.03&89.93&90.95&\textbf{91.29}
&77.67&86.07&83.16&85.03&\textbf{86.00}&85.91
&64.92&68.23&73.03&74.44&80.66&\textbf{82.09}\\
\hline

\multicolumn{4}{c|}{\textbf{Average}}
&79.06&80.70&82.29&83.63&86.81&\textbf{88.36}
&66.67&71.27&71.36&74.28&78.65&\textbf{79.86}
&48.41&52.87&55.22&56.70&64.11&\textbf{66.67}\\

\bottomrule
\end{tabular}
\end{adjustbox}

\label{table:detailed_improvment}
\end{table*}    
\begin{table*}[h]
\centering
\Huge
\caption{Ablation study to evaluate the effectiveness of our proposed pre-training method with synthetic data. The baselines RFNet, RobustSeg, and UHVED are denoted as RF, RS, and UH, respectively, due to space limitations. The mixup data augmentation method\cite{mixup} is represented by `MU', while AEM (Asymmetry Error Map) is the proposed pre-training method based on synthetic data using AEMs. All the experimental results were obtained using the authors' codes. The available and missing modalities are denoted by \textbf{$\bullet$} and \textbf{$\circ$}, respectively. The \textbf{bold} font shows the best.}
\resizebox{1\linewidth}{!}{
\begin{tabular}{cccc|ccc|ccc|ccc|ccc|ccc|ccc|ccc|ccc|ccc}
 \toprule
\multicolumn{4}{c|}{ } & \multicolumn{27}{c}{ \textbf{Dice Score(\%)}} \\ \cline{5-31} 
\multicolumn{4}{c|}{\multirow{-2}{*}{ \textbf{Modalities}}} & \multicolumn{9}{|c|}{ \textbf{Whole}} & \multicolumn{9}{c|}{ \textbf{Core}} & \multicolumn{9}{c}{ \textbf{Enhancing}} \\ \midrule
 \textbf{F} &  \textbf{T1} &  \textbf{T1ce} &  \textbf{T2} &  \textbf{RF} &  \textbf{+MU} &  \textbf{+AEM} &  \textbf{RS} &  \textbf{+MU} &  \textbf{+AEM} &  \textbf{UH} &  \textbf{+MU} &  \textbf{+AEM} &  \textbf{RF} &  \textbf{+MU} &  \textbf{+AEM} &  \textbf{RS} &  \textbf{+MU} &  \textbf{+AEM} &  \textbf{UH} &  \textbf{+MU} &  \textbf{+AEM} &  \textbf{RF} &  \textbf{+MU} &  \textbf{+AEM} &  \textbf{RS} &  \textbf{+MU} &  \textbf{+AEM} &  \textbf{UH} &  \textbf{+MU} &  \textbf{+AEM} \\ \midrule
 \textbf{$\circ$}&\textbf{$\circ$}&\textbf{$\circ$}&\textbf{$\bullet$}
&86.18&85.19&\textbf{86.23}&81.29&82.93&\textbf{82.99}&76.57&\textbf{78.67}&78.60
&71.14&\textbf{71.99}&70.68&60.97&61.84&\textbf{62.12}&54.96&57.30&\textbf{62.45}
&47.53&49.65&\textbf{49.98}&38.25&39.00&\textbf{39.19}&32.05&32.96&\textbf{37.81}\\
\textbf{$\circ$}&\textbf{$\circ$}&\textbf{$\bullet$}&\textbf{$\circ$}
&77.68&76.69&\textbf{78.97}&\textbf{67.31}&66.78&66.68&66.48&\textbf{66.53}&64.90
&82.90&81.86&\textbf{83.30}&73.26&74.70&\textbf{74.93}&71.19&72.65&\textbf{73.83}
&77.01&73.54&\textbf{78.33}&65.64&65.37&\textbf{66.09}&60.48&59.97&\textbf{63.00}\\
\textbf{$\circ$}&\textbf{$\bullet$}&\textbf{$\circ$}&\textbf{$\circ$}
&76.00&76.11&\textbf{78.44}&60.25&64.16&\textbf{64.29}&59.56&60.18&\textbf{62.00}
&62.61&65.67&\textbf{66.57}&43.74&49.82&\textbf{50.20}&44.63&44.15&\textbf{49.19}
&37.70&40.31&\textbf{40.91}&21.23&24.02&\textbf{24.22}&11.90&19.65&\textbf{22.31}\\
\textbf{$\bullet$}&\textbf{$\circ$}&\textbf{$\circ$}&\textbf{$\circ$}
&86.79&87.25&\textbf{87.59}&\textbf{82.55}&82.10&82.16&77.18&\textbf{80.34}&79.60
&69.42&69.36&\textbf{69.98}&55.94&58.43&\textbf{58.69}&52.02&\textbf{57.03}&55.17
&43.36&43.69&\textbf{45.35}&31.07&32.90&\textbf{33.05}&30.32&\textbf{35.73}&25.08\\
\textbf{$\circ$}&\textbf{$\circ$}&\textbf{$\bullet$}&\textbf{$\bullet$}
&88.19&87.82&\textbf{88.56}&85.00&85.62&\textbf{85.73}&82.30&\textbf{83.73}&82.20
&\textbf{85.82}&84.52&85.24&81.88&82.67&\textbf{83.05}&75.35&79.77&\textbf{80.58}
&78.16&75.56&\textbf{80.24}&72.08&73.60&\textbf{73.81}&65.54&65.54&\textbf{68.40}\\
\textbf{$\circ$}&\textbf{$\bullet$}&\textbf{$\bullet$}&\textbf{$\circ$}
&81.02&80.49&\textbf{82.04}&73.16&73.48&\textbf{73.57}&71.65&\textbf{71.75}&70.79
&\textbf{83.89}&82.82&83.33&79.47&79.59&\textbf{79.85}&76.37&77.30&\textbf{78.08}
&77.79&78.68&\textbf{79.56}&73.01&73.17&\textbf{73.35}&65.00&64.14&\textbf{67.04}\\
\textbf{$\bullet$}&\textbf{$\bullet$}&\textbf{$\circ$}&\textbf{$\circ$}
&89.15&89.50&\textbf{90.24}&\textbf{86.54}&86.48&\textbf{86.54}&82.56&\textbf{84.49}&83.24
&72.41&72.97&\textbf{75.12}&63.48&65.82&\textbf{66.11}&60.50&62.41&\textbf{62.44}
&47.22&\textbf{48.76}&47.51&37.35&38.06&\textbf{38.32}&32.83&\textbf{37.38}&32.84\\
\textbf{$\circ$}&\textbf{$\bullet$}&\textbf{$\circ$}&\textbf{$\bullet$}
&87.73&87.04&\textbf{88.32}&84.78&85.37&\textbf{85.44}&81.64&\textbf{82.68}&81.72
&73.19&\textbf{73.41}&72.97&65.41&67.12&\textbf{67.33}&60.77&61.67&\textbf{65.40}
&50.81&\textbf{54.22}&50.40&40.46&40.86&\textbf{40.98}&32.55&34.68&\textbf{39.23}\\
\textbf{$\bullet$}&\textbf{$\circ$}&\textbf{$\circ$}&\textbf{$\bullet$}
&89.26&89.19&\textbf{90.37}&87.27&88.16&\textbf{88.23}&82.19&\textbf{84.24}&83.64
&74.33&74.22&\textbf{74.37}&66.30&67.70&\textbf{67.99}&59.28&62.66&\textbf{65.19}
&53.67&53.80&\textbf{54.20}&42.77&42.60&\textbf{42.80}&35.62&38.14&\textbf{40.80}\\
\textbf{$\bullet$}&\textbf{$\circ$}&\textbf{$\bullet$}&\textbf{$\circ$}
&89.67&88.99&\textbf{90.44}&86.60&86.54&\textbf{86.64}&82.23&\textbf{85.84}&83.05
&85.50&85.10&\textbf{85.52}&80.80&81.42&\textbf{81.74}&73.61&77.11&\textbf{77.58}
&76.77&75.32&\textbf{79.28}&\textbf{70.54}&69.41&69.64&63.07&62.96&\textbf{67.01}\\
\textbf{$\bullet$}&\textbf{$\bullet$}&\textbf{$\bullet$}&\textbf{$\circ$}
&90.24&90.00&\textbf{91.01}&87.51&87.74&\textbf{87.81}&84.42&\textbf{87.07}&84.39
&\textbf{85.65}&84.60&85.08&82.53&82.47&\textbf{82.77}&76.97&78.73&\textbf{79.02}
&80.00&81.13&\textbf{82.40}&73.02&73.13&\textbf{73.27}&64.75&64.39&\textbf{67.84}\\
\textbf{$\bullet$}&\textbf{$\bullet$}&\textbf{$\circ$}&\textbf{$\bullet$}
&90.26&90.15&\textbf{90.92}&88.69&88.80&\textbf{88.91}&84.56&\textbf{86.35}&85.28
&74.88&\textbf{75.57}&75.32&68.33&70.00&\textbf{70.24}&63.38&65.24&\textbf{68.29}
&52.38&\textbf{54.16}&52.44&44.42&43.20&\textbf{43.38}&37.07&39.23&\textbf{41.25}\\
\textbf{$\bullet$}&\textbf{$\circ$}&\textbf{$\bullet$}&\textbf{$\bullet$}
&90.30&90.24&\textbf{91.27}&88.49&88.88&\textbf{88.98}&84.72&\textbf{87.52}&85.41
&\textbf{85.88}&85.42&85.39&82.20&82.71&\textbf{83.02}&75.62&78.80&\textbf{79.79}
&79.23&77.03&\textbf{79.61}&71.41&71.31&\textbf{71.51}&63.75&63.81&\textbf{67.42}\\
\textbf{$\circ$}&\textbf{$\bullet$}&\textbf{$\bullet$}&\textbf{$\bullet$}
&88.72&88.54&\textbf{89.19}&85.81&86.31&\textbf{86.43}&84.04&\textbf{84.90}&83.31
&\textbf{86.06}&84.18&83.95&82.86&83.34&\textbf{83.66}&77.68&80.53&\textbf{81.08}
&79.31&78.33&\textbf{81.65}&73.96&75.23&\textbf{75.41}&66.31&66.59&\textbf{69.01}\\
\textbf{$\bullet$}&\textbf{$\bullet$}&\textbf{$\bullet$}&\textbf{$\bullet$}
&90.95&90.63&\textbf{91.49}&89.03&89.00&\textbf{89.08}&85.84&\textbf{88.16}&85.97
&\textbf{86.00}&85.16&85.15&83.16&83.21&\textbf{83.50}&77.67&79.60&\textbf{80.56}
&80.66&80.45&\textbf{82.73}&73.03&73.50&\textbf{73.70}&64.92&64.85&\textbf{67.89}\\ \hline

\multicolumn{4}{c|}{\textbf{Average}}
&86.81&86.52&\textbf{87.67}&82.29&82.81&\textbf{82.90}&79.06&\textbf{80.83}&79.60
&78.65&78.46&\textbf{78.80}&71.36&72.72&\textbf{73.01}&66.67&69.00&\textbf{70.57}
&64.11&64.31&\textbf{65.64}&55.22&55.69&\textbf{55.91}&48.41&50.00&\textbf{51.80}\\ \bottomrule
\end{tabular}
}

\label{table:augmentation}
\end{table*}
\begin{table*}[h]
\Huge
\centering
\caption{Ablation study of the effectiveness of our proposed pre-training method on RFNet as the baseline by varying the number of Asymmetric Error Maps (AEMs) used for pre-training. The number of synthetic training samples used for pre-training was denoted as `AEM$\times$x', where `x' indicates the size of the augmented dataset, x times the original dataset. By increasing the size of the augmented dataset, we observed an overall improvement in the performance of the network, demonstrating the effectiveness of our pre-training method in preparing the network to recognize abnormal tumor intensities across different samples.
}
\renewcommand\arraystretch{1} 
\resizebox{\linewidth}{!}{
\begin{tabular}{cccc|cccc|cccc|cccc}
\toprule
\multicolumn{4}{c|}{}                                                                                                                                  & \multicolumn{12}{c}{\textbf{Dice score(\%)}}\\ \cline{5-16} 
\multicolumn{4}{c|}{\multirow{-2}{*}{\textbf{Modality}}}                                                                                               & \multicolumn{4}{c|}{\textbf{Whole}}                                                                                                                                                                        & \multicolumn{4}{c|}{\textbf{Core}}                                                                                                                                                                         & \multicolumn{4}{c}{\textbf{Enhancing}}                                                                                                                                                                     \\
\midrule
{\textbf{F}}&{\textbf{T1}}&{\textbf{T1ce}}&{\textbf{T2}}&{\textbf{Baseline}}&{\textbf{AEM$\times$2}}&{\textbf{AEM$\times$4}}&{\textbf{AEM$\times$8}}&{\textbf{Baseline}}&{\textbf{AEM$\times$2}}&{\textbf{AEM$\times$4}}&{\textbf{AEM$\times$8}}&{\textbf{Baseline}}&{\textbf{AEM$\times$2}}&{\textbf{AEM$\times$4}}&{\textbf{AEM$\times$8}}\\
\midrule
\textbf{$\circ$}&\textbf{$\circ$}&\textbf{$\circ$}&\textbf{$\bullet$}
&86.18&86.24&\textbf{86.76}&86.23
&71.14&\textbf{71.99}&71.33&70.68
&47.53&\textbf{51.74}&49.99&49.98\\
\textbf{$\circ$}&\textbf{$\circ$}&\textbf{$\bullet$}&\textbf{$\circ$}
&77.68&79.07&\textbf{79.53}&78.97
&82.90&82.88&82.64&\textbf{83.30}
&77.01&76.00&77.26&\textbf{78.33}\\
\textbf{$\circ$}&\textbf{$\bullet$}&\textbf{$\circ$}&\textbf{$\circ$}
&76.00&78.70&\textbf{79.94}&78.44
&62.61&\textbf{67.36}&67.19&66.57
&37.70&40.86&\textbf{41.53}&40.91\\
\textbf{$\bullet$}&\textbf{$\circ$}&\textbf{$\circ$}&\textbf{$\circ$}
&86.79&\textbf{88.05}&87.66&87.59
&69.42&69.27&\textbf{70.62}&69.98
&43.36&41.94&44.36&\textbf{45.35}\\
\textbf{$\circ$}&\textbf{$\circ$}&\textbf{$\bullet$}&\textbf{$\bullet$}
&88.19&88.34&88.47&\textbf{88.56}
&\textbf{85.82}&84.87&84.80&85.24
&78.16&78.87&79.96&\textbf{80.24}\\
\textbf{$\circ$}&\textbf{$\bullet$}&\textbf{$\bullet$}&\textbf{$\circ$}
&81.02&82.08&\textbf{82.49}&82.04
&\textbf{83.89}&83.01&83.24&83.33
&77.79&76.64&\textbf{81.14}&79.56\\
\textbf{$\bullet$}&\textbf{$\bullet$}&\textbf{$\circ$}&\textbf{$\circ$}
&89.15&90.01&90.06&\textbf{90.24}
&72.41&73.18&73.45&\textbf{75.12}
&47.22&47.30&\textbf{49.96}&47.51\\
\textbf{$\circ$}&\textbf{$\bullet$}&\textbf{$\circ$}&\textbf{$\bullet$}
&87.73&88.03&88.26&\textbf{88.32}
&\textbf{73.19}&72.95&72.98&72.97
&50.81&50.21&\textbf{51.09}&50.40\\
\textbf{$\bullet$}&\textbf{$\circ$}&\textbf{$\circ$}&\textbf{$\bullet$}
&89.26&90.18&89.96&\textbf{90.37}
&74.33&73.86&\textbf{74.91}&74.37
&53.67&50.37&51.16&\textbf{54.20}\\
\textbf{$\bullet$}&\textbf{$\circ$}&\textbf{$\bullet$}&\textbf{$\circ$}
&89.67&90.13&90.00&\textbf{90.44}
&85.50&\textbf{85.61}&85.49&85.52
&76.77&\textbf{79.74}&78.93&79.28\\
\textbf{$\bullet$}&\textbf{$\bullet$}&\textbf{$\bullet$}&\textbf{$\circ$}
&90.24&90.72&90.77&\textbf{91.01}
&\textbf{85.65}&85.38&85.44&85.08
&80.00&82.23&82.14&\textbf{82.40}\\
\textbf{$\bullet$}&\textbf{$\bullet$}&\textbf{$\circ$}&\textbf{$\bullet$}
&90.26&90.50&90.57&\textbf{90.92}
&74.88&74.75&75.03&\textbf{75.32}
&52.38&50.59&\textbf{52.47}&52.44\\
\textbf{$\bullet$}&\textbf{$\circ$}&\textbf{$\bullet$}&\textbf{$\bullet$}
&90.30&\textbf{90.95}&90.81&89.19
&85.88&85.53&\textbf{85.94}&85.39
&79.23&79.86&\textbf{80.19}&79.61\\
\textbf{$\circ$}&\textbf{$\bullet$}&\textbf{$\bullet$}&\textbf{$\bullet$}
&88.72&88.95&89.08&\textbf{89.19}
&\textbf{86.06}&84.35&84.34&83.95
&79.31&80.48&\textbf{81.95}&81.65\\
\textbf{$\bullet$}&\textbf{$\bullet$}&\textbf{$\bullet$}&\textbf{$\bullet$}
&90.95&91.06&91.08&\textbf{91.49}
&\textbf{86.00}&85.51&84.80&85.15
&80.66&81.36&82.07&\textbf{82.73}\\
\midrule
\multicolumn{4}{c|}{\textbf{Average}}
&86.81&87.51&87.61&\textbf{87.67}
&78.65&78.75&\textbf{78.81}&78.80
&64.11&64.54&65.61&\textbf{65.64}\\
\bottomrule
\end{tabular}
}
\label{table:Different_dataset_size}
\end{table*}   
\begin{table}[h]
\caption{The results of finetuning after the pre-training on augmented data. All results are reported with BRATS2020.}
\centering
\resizebox{0.6\textwidth}{!}{
\begin{tabular}{l|l|l|l} 
\hline
\textbf{Method}        & \textbf{Whole}        & \textbf{Core}         & \textbf{Enhancing}     \\ 
\hline
Baseline        & 86.81        & 78.65        & 64.11         \\
Synthetic    & 85.38$_{\color{blue}{-1.43}}$ & 74.98$_{\color{blue}{-3.67}}$ & 58.95$_{\color{blue}{-5.16}}$   \\
Real (Ours) & 87.67$_{\color{red}{+0.86}}$  & 78.80$_{\color{red}{+0.15}}$  & 65.64$_{\color{red}{+1.53}}$  \\ 
\hline
\end{tabular}
}
\label{table:why_finetune}
\end{table}
\begin{table}[h]
\centering
\caption{
Comparison of our KD-based post-training with the `One-to-many' model and the `One-to-one' model. Our KD-based post-training achieves the best performance. All results are reported with Dice Score (\%).
}
\Huge
\resizebox{0.9\linewidth}{!}{
\begin{tabular}{c|c|c|c|c|c|c|c|c|c}
\toprule
\multirow{2}{*}{\textbf{Method}} & \multicolumn{3}{c|}{\textbf{Whole}} & \multicolumn{3}{c|}{\textbf{Core}} & \multicolumn{3}{c}{\textbf{Enhancing}} \\ \cline{2-10}
 & \textbf{T1} & \textbf{T1T2} & \textbf{T1T2F} & \textbf{T1} & \textbf{T1T2} & \textbf{T1T2F} & \textbf{T1} & \textbf{T1T2} & \textbf{T1T2F} \\ \midrule
\textbf{One-to-Many} & 76.00 & 87.73 & 90.26 & 62.61 & 73.19 & 74.88 & 37.70 & 50.81 & 52.38 \\
\textbf{One-to-One} & 79.58 & 88.27 & 90.60 & 64.37 & 72.64 & 75.40 & 38.65 & 49.89 & 52.74 \\ \midrule
\textbf{KD-based} & \textbf{81.35} & \textbf{88.83} & \textbf{90.87} & \textbf{69.95} & \textbf{74.04} & \textbf{75.51} & \textbf{42.29} & \textbf{53.92} & \textbf{52.79} \\ \bottomrule
\end{tabular}
}

\label{Table:one-many}
\end{table}               
\begin{table}[h]
\caption{Ablation study of investigating the impact of increasing the number of training iterations of the Knowledge Distillation (KD)-based post-training method. The notation KD-$N$ represents that the model was post-trained with $N$ epochs using our proposed KD approach. The results are reported in terms of Dice Score (\%) on the BRATS2020 dataset.}
\centering
\resizebox{0.7\textwidth}{!}{

\begin{tabular}{l|c|c|c} 
\hline
\textbf{Method}        & \textbf{Whole}        & \textbf{Core}         & \textbf{Enhancing}     \\ 
\hline
UHVED+AEM       & 79.61 & 70.58 & 51.80  \\ 
UHVED+AEM+KD300 & 79.95 & 70.56 & 52.09  \\ 
UHVED+AEM+KD600 & 80.20 & 70.91 & 52.50  \\ 
UHVED+AEM+KD900 & \textbf{80.70} & \textbf{71.27} & \textbf{52.87}  \\ 
\hline
RobustSeg+AEM       & 82.90 & 73.01 & 55.91   \\ 
RobustSeg+AEM+KD300 & 83.25 & 73.22 & 55.87   \\ 
RobustSeg+AEM+KD600 & 83.54 & 73.80 & 56.17   \\ 
RobustSeg+AEM+KD900 & \textbf{83.63} & \textbf{74.28} & \textbf{56.70}   \\ 
\hline
RFNet+AEM        & 87.67& 78.80  & 65.64  \\
RFNet+AEM+KD300  & 87.71& 79.31  & 65.61  \\
RFNet+AEM+KD600  & 87.99& 78.87  & 65.99  \\
RFNet+AEM+KD900  & \textbf{88.36} & \textbf{79.86}  & \textbf{66.67}  \\
\hline
\end{tabular}

}
\label{table:ablation_KD}
\end{table}

\begin{figure}[htb!]
  \centering
    \includegraphics[width=0.8\linewidth]{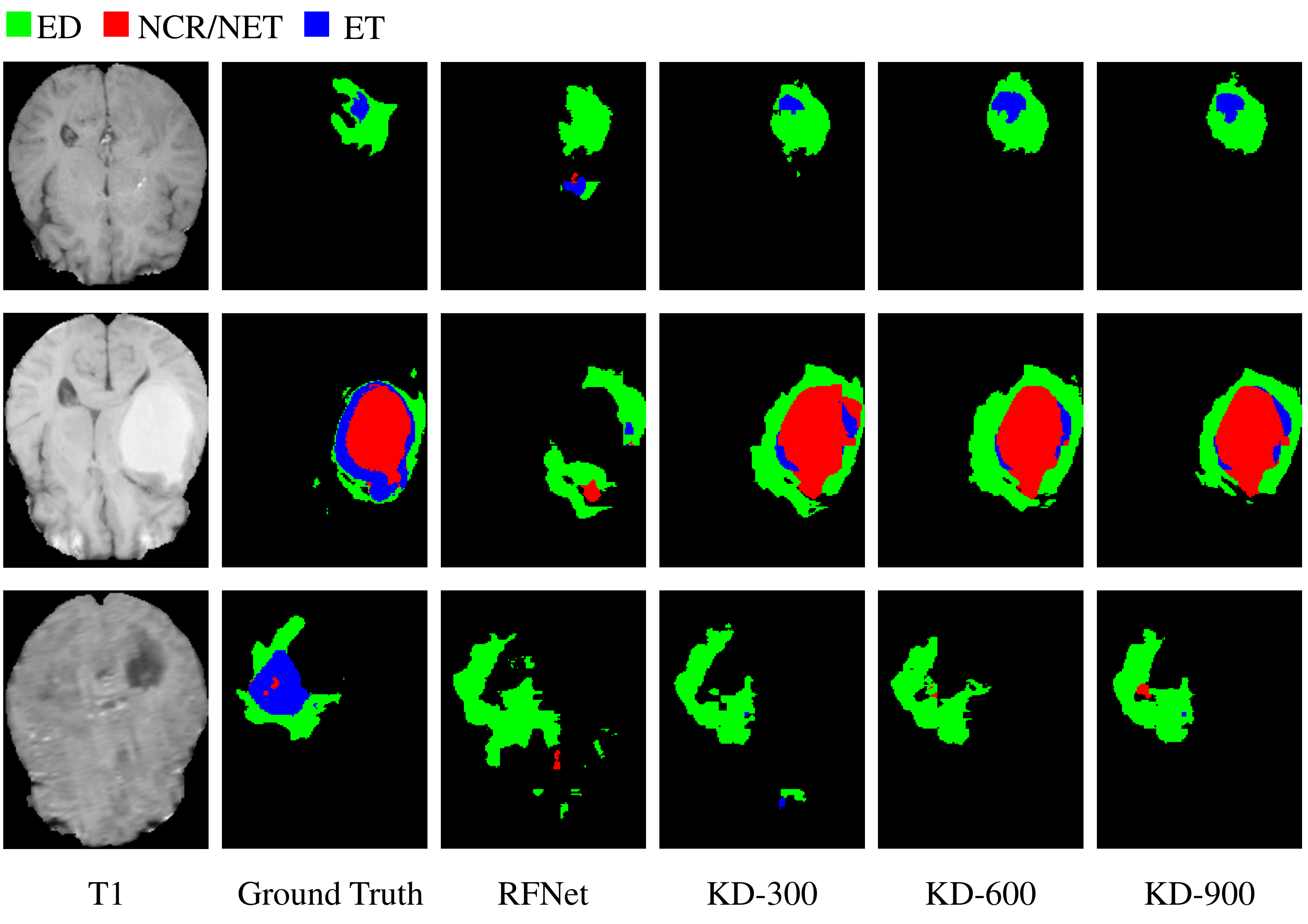}
    \caption{
    Visualizations of the prediction of different iterations for the KD-based post-training, using only the T1 modality as input, are presented. At the early stage of training, our method may show similarities to RFNet\cite{ding2021rfnet} (Baseline). However, as the number of training iterations increases, the performance of our method progressively improves.
    }
    \label{Figure:ablation_KD}
\end{figure}   

\textbf{Visualization of the synthetic samples}
Fig.~\ref{Figure: Additional_synthetic_samples} presents a visual representation of the synthetic samples generated using our Asymmetric Error Maps (AEM). To create the synthetic samples, we integrated the tumor part of sample $X_B$ into sample $X_A$, resulting in sample $X_{AB}$. The corresponding ground truth, $Y_{AB}$, was generated by merging $Y_A$ and $Y_B$. To better illustrate the process, we have included an example of each modality (Flair, T1ce, T1, and T2) in the figure, arranged in top-to-bottom order.

\subsection{Knowledge Distillation-based Post-training}
We propose a post-training stage based on Knowledge Distillation (KD) to handle missing modalities, as illustrated in Fig.~\ref{Figure-knowledge}(c). The approach involves three steps:
\begin{enumerate}[Step 1:]
\item \textbf{Weight Initialization.}: The weights of the student network are initialized with the same values as the teacher network.
\item \textbf{Knowledge Distillation.}: During this stage, only the student network is fine-tuned while the teacher network remains frozen. One modality is randomly removed from the inputs to the teacher network to form the inputs for the student network. The intermediate features and intermediate segmentation predictions of the teacher network serve as extra supervision for the student network.
\item \textbf{Update teacher model.}: After every $k$ ($k=5$) epochs of training, the teacher network is updated by replacing it with the latest generated student network.
\end{enumerate}
These steps are repeated until the student network converges. With the teacher network receiving more modalities than the student, the student network gradually learns to recover missing information in its predictions.

As discussed in Step 2, the goal of the post-training stage is to train the student model to produce intermediate features and segmentation predictions that are similar to those generated by the teacher model. To achieve this, the total loss function $L_{post}$ in the knowledge distillation (KD)-based post-training process is composed of two parts. The first part, denoted as $L_{kd}$, measures the distance between the intermediate features generated by the student and teacher models. The second part, denoted as $L_{seg}$, is the segmentation loss used in the original model~\cite{ding2021rfnet,chen2019robust,dorent2019hetero}. By combining these two losses, the post-training stage aims to optimize the student model to generate accurate segmentation predictions while also mimicking the intermediate features produced by the teacher model.

Specifically, $L_{kd}$ is formulated as the mean squared error between the teacher feature $f_t$ and student feature $f_s$:
\begin{equation}
L_{kd}= \text{MSE}(f_t, f_s).
\end{equation}
The overall post-training loss is then given by the sum of $L_{kd}$ and $L_{seg}$:
\begin{equation}
L_{post} = L_{kd} + L_{seg}.
\end{equation}

\section{Experiments}\label{experiment}
\subsection{Experimental Settings}
\noindent \textbf{Datasets and Evaluation Metric.} 
For a fair comparison with previous incomplete multi-modal brain tumor segmentation methods ~\cite{ding2021rfnet,dorent2019hetero,chen2019robust}, we evaluated our proposed method on three Multimodal Brain Tumor Segmentation Challenge datasets: BRATS 2020, BRATS 2018, and BRATS 2015. In incomplete multi-modal brain tumor segmentation, the goal is to identify the regions of the tumor in multi-modal MRI images, including the whole tumor, tumor core, and enhancing tumor.

To consistent with previous methods ~\cite{ding2021rfnet,dorent2019hetero,chen2019robust}, we used the Dice score ~\cite{dice1945measures} as the validation metric for evaluating brain tumor segmentation performance. Following previous methods, we divided the 369 BRATS2020 samples into 219 training samples, 50 validation samples, and 100 test samples. Similarly, we divided BRATS2018 into 199 training samples, 29 validation samples, and 57 test samples, while BRATS2015 was split into 242, 12, and 20 samples for training, validation, and testing, respectively.

\noindent \textbf{Implementation Details.} 
Following previous methods~\cite{ding2021rfnet,dorent2019hetero,chen2019robust}, we removed non-meaningful black background areas from the MRI images and normalized them with zero mean and unit variance. During training, we randomly cropped the MRI images to 80 $\times$ 80 $\times$ 80, and conducted further pre-processing operations, including random rotations, intensity shifts, and random flipping, for data augmentation. Unless otherwise stated, we use RFNet ~\cite{ding2021rfnet} as the baseline model to combine with our pre-training and post-training strategy. We train the network for 300 epochs using Adam optimizer ~\cite{kingma2014adam} with a poly learning rate strategy, with the initial learning rate set to 2e-4.

\subsection{Comparison with State-of-the-art}
\noindent \textbf{BRATS2020.} To evaluate the effectiveness of the proposed approach, we compare it to several state-of-the-art methods, such as HeMIS~\cite{have2016}, UHVED~\cite{dorent2019hetero}, RobustSeg~\cite{chen2019robust} and RFNet~\cite{ding2021rfnet}. Table~\ref{table:soa_2020} shows a comparison of our method with the previous state-of-the-art methods on the BRATS2020 dataset. Our proposed method outperforms all other methods in all three tumor categories, with an average improvement of 2.56\% in enhancing tumor segmentation compared to the best performing method, RFNet~\cite{ding2021rfnet}. For single-modality performance (T2/T1ce/T1/F), our approach demonstrates 3.95\%, 2.31\%, 4.59\% and 0.87\% improvements in accurately segmenting the enhancing tumors. For whole tumor and core segmentation, our method achieves 1.38\%, 3.77\%, 5.35\%, and 1.77\% and 2.34\%, 1.24\%, 7.34\%, and 2.49\% improvements, respectively. 
Furthermore, Fig.~\ref{Figure: different_method_prediction} visually compares the segmentation maps of our proposed method with those of other state-of-the-art methods, utilizing all four modalities. The figure shows that our method accurately located the tumors and covered the majority of the tumor regions, while the predictions from previous methods resulted in incomplete or incorrect segmentation.

\noindent \textbf{BRATS2018.}
Table~\ref{table:soa_2018} presents the evaluation results of our proposed method on the BRATS2018 dataset, which demonstrate consistent and superior performance compared to other state-of-the-art methods, as previously observed on the BRATS2020 dataset. On average, our approach outperforms the best performing method RFNet by 4.55\% in terms of enhancing tumor segmentation accuracy. Moreover, our method consistently achieves significant improvements over RFNet in all three tumor categories (whole tumor, tumor core, and enhancing tumor) when dealing with single-modality segmentation. These results validate the efficacy of our proposed approach in enhancing the accuracy of brain tumor segmentation.

\noindent \textbf{BRATS2015.} 
Table~\ref{table:soa_2015} shows the evaluation results of the proposed method on the BRATS2015 dataset, indicating our method's effectiveness and consistent improvements compared to other state-of-the-art methods. The proposed method demonstrates significant advancements in segmentation accuracy, similar to the conclusion achieved on the BRATS2020 and BRATS2018 datasets. These results reinforce the robustness and effectiveness of the proposed method in handling various multimodal brain tumor segmentation tasks.

\subsection{Ablation Studies}

\subsubsection{The effectiveness of each component of our method} 
To validate the efficacy of each component of our method, we integrated them with three state-of-the-art (SOTA) methods: UHVED~\cite{dorent2019hetero}, RobustSeg~\cite{chen2019robust}, and RFNet~\cite{ding2021rfnet}. The results of our proposed Asymmetric Error Maps (AEM) pre-training and (Knowledge Distill) KD-based post-training are presented in Table~\ref{table:method_with_ours}, which demonstrates a significant improvement over all the baselines. Detailed performance analysis in Table~\ref{table:detailed_improvment} shows that our proposed method achieved significant performance improvements on almost all metrics for the baseline model, particularly under the condition of single-modality input. Specifically, when only T1 is given as input, our method demonstrated improvements of 12.34\%, 5.20\%, and 4.59\% for accurately segmenting the enhancing tumors, and 2.13\%, 7.81\%, and 5.35\% for whole tumor segmentation, and 5.87\%, 10.00\%, and 7.31\% for core segmentation, respectively.

In addition, as shown in Fig.~\ref{Figure: methods_with_ours}, we visually validate the effectiveness of our proposed methods for brain tumor segmentation with different baselines. The figure clearly demonstrates the superiority of our method in generating high-quality predictions compared to the different baseline models.

\subsubsection{The effectiveness of the pre-training}
\noindent \textbf{Asymmetric Error Maps V.S. MixUp}
As depicted in Table~\ref{table:augmentation}, our proposed pre-training method with Asymmetric Error Maps (AEM) was validated with several existing methods, such as UHVED~\cite{dorent2019hetero}, RobustSeg~\cite{chen2019robust}, and RFNet~\cite{ding2021rfnet}.  In each case, the models were pre-trained using our method and then fine-tuned with real training samples. Our validation results indicate that the proposed pre-training method outperforms the baseline. We also compared our data synthesis approach to the popular data augmentation technique mixup~\cite{mixup}. To conduct this comparison, we generated synthesized data using mixup and pre-trained the models before fine-tuning the model with real training samples. As shown in Table~\ref{table:augmentation} our proposed pre-training method with AEM is more effective than the mixup method in most cases. All experiments were conducted using the BRATS2020 dataset.

\noindent \textbf{The effectiveness of the number of synthetic data.}
Table~\ref{table:Different_dataset_size} illustrates the impact of increasing the number of Asymmetric Error Maps (AEM) used as synthetic data in the pre-training phase on the overall performance of the baselines. The results demonstrate an improvement in prediction accuracy when applying the AEM pre-training strategy. However, further increasing the number of pre-training synthetic data does not always lead to improvement. For instance, when the synthetic data size is eight times the real data size, the performance is similar to when the synthetic data size is four times the real data size. We hypothesize that this is because when the synthetic data size reaches four times the real data size, the diversity of the training set may have already achieved saturation, and further increases in synthetic data may not be as effective in improving performance. In summary, collecting a sufficient number of AEM is crucial for effective pre-training, and doing so can ensure the optimal performance of the model.

\noindent \textbf{The effectiveness of the finetuning on real data.}
Directly fine-tuning both synthetic and real data may lead to bias towards the synthetic data distribution due to its larger size compared to the real data. To mitigate this bias, we fine-tune our model exclusively on the real data. The effectiveness of this approach is demonstrated in Table~\ref{table:why_finetune}, which shows a significant improvement over the baseline. In contrast, direct fine-tuning of the combined synthetic and real data may result in performance degradation.

\subsubsection{The effectiveness of the post-training}
\noindent \textbf{One-to-One vs. One-to-Many vs. KD-based Post-training.}
To further demonstrate the effectiveness of our proposed KD-based post-training approach, we conducted a comparison with the ``One-to-Many" and ``One-to-One" variants. The former trains a single model to handle all combinations of modalities, while the latter trains individual models for each specific scenario. It is expected that the one-to-one model would perform better than the one-to-many model as it is trained without irrelevant information. As shown in Table~\ref{Table:one-many}, our assumption was confirmed, with the ``one-to-one" model outperforming the ``one-to-many" model. However, the one-to-one method faces a trade-off between training complexity and accuracy.

In contrast, our KD-based post-training strategy trains a robust student model that can handle all scenarios with a unified model, without increasing the training complexity. The results demonstrate that our approach outperforms both the one-to-one and one-to-many methods, achieving the best performance in handling different input modalities.

\noindent \textbf{The effectiveness of different iterations for KD-based post-training.}
Table~\ref{table:ablation_KD} presents the results of our KD-based post-training approach combined with three existing methods: UHVED~\cite{dorent2019hetero}, RobustSeg~\cite{chen2019robust}, and RFNet~\cite{ding2021rfnet}. In this ablation experiment, we applied our AEM pre-training to these models as the baselines, followed by the KD-based post-training with different finetuning epochs. The results indicate that our proposed post-training method outperforms the baselines significantly, and the performance continues to improve as the iterations of the KD-based post-training increase. 

Furthermore, to better understand how our KD-based post-training approach improves the baseline segmentation with different finetuning epochs, we visualize the predicted segmentation results as the number of training iterations increases. Fig.~\ref{Figure:ablation_KD} shows that our method predicts more accurate masks as the number of iterations increases, providing insights into the benefits of our approach.

\subsection{Visualization of the segmentation results}
\noindent  \textbf{Increasing the modality information.}
Fig.~\ref{Figure: different_modal_prediction} highlights the impact of incorporating additional modality information on the segmentation of our predictions. As the number of modalities increases, our method can produce increasingly precise segmentation masks, indicating its enhanced capability to identify and differentiate various tissue structures. The results demonstrate the effectiveness of our approach and its capacity to generate high-quality predictions by leveraging additional modality information.

\section{Conclusion}
In summary, we propose a novel approach for segmenting brain tumors from multi-modal MRI images that involves both synthetic data-based pre-training and knowledge distillation-based post-training. Our pre-training method utilizes the asymmetrical nature of abnormal brain MRI images to synthesize training samples by adding tumor intensities to healthy brain anatomy. This prepares the model to recognize abnormal intensities across various anatomical regions before the actual training. The post-training method further improves the model's performance by distilling knowledge from intermediate features and predictions obtained from fully-modal inputs. Our approach outperforms existing methods and achieves state-of-the-art results on three benchmark datasets: BRATS2020, BRATS2018, and BRATS2015.

\bibliographystyle{elsarticle-num}
\bibliography{egbib}
\end{document}